%% file: arXiv.tex
\documentclass[sigconf]{acmart}

\usepackage{booktabs}
\usepackage{threeparttable} 
\usepackage{multirow}
\usepackage{xcolor} 

\usepackage{soul} 
\usepackage{xspace}
\definecolor{LightOrange}{RGB}{255, 224, 178}
\definecolor{LightPurple}{RGB}{235, 222, 240}
\definecolor{LightBrown}{RGB}{222, 184, 135}
\definecolor{LightTeal}{RGB}{175, 238, 238}
\definecolor{LightCoral}{RGB}{240, 128, 128}
\definecolor{LightLavender}{RGB}{230, 230, 250}

\usepackage{tabularx}
\usepackage{array}
\usepackage{geometry}
\usepackage{makecell} 
\usepackage{lineno}

\usepackage[normalem]{ulem}
\useunder{\uline}{\ul}{}

\usepackage{tabularx}
\usepackage{tcolorbox}

\usepackage{amsmath}
\usepackage{dsfont}
\usepackage{xfrac}
\usepackage{amsfonts}
\usepackage{pifont}
\usepackage{mdwlist}   
\usepackage{paralist}  
\usepackage{enumitem}

\usepackage{epigraph}
\usepackage{mdframed}
\usepackage{makecell}
\usepackage{ulem} %
\usepackage{pbox} 

\colorlet{deepgreen}{green!60!black}
\colorlet{deepred}{red!60!brown}
\usepackage{tikz}

\setcopyright{acmlicensed}
\copyrightyear{2025}
\acmYear{2025}
\acmSubmissionID{101}

\acmBooktitle{CCS '25: October 13–17, 2025, Taipei, Taiwan}
\acmISBN{978-1-4503-XXXX-X/18/06}

\input{dfn}

\usepackage[font=small,skip=4pt]{caption} %
\begin{document}

\title[\wholeframe]{Rethinking Machine Unlearning in Image Generation Models}

\author{Renyang Liu}
\authornote{Equal contribution.}
\affiliation{%
  \institution{Institute of Data Science,
  National University of Singapore}
  \city{Singapore}
  \country{Singapore}
  }
\email{ryliu@nus.edu.sg}

\author{Wenjie Feng}
\authornotemark[1]
\authornote{Corresponding author; Project lead.}
\affiliation{%
  \institution{
       School of Artificial Intelligence and Data Science,
       University of Science and Technology of China
  }
  \city{Hefei}
  \state{Anhui}
  \country{China}
}
\email{fengwenjie@ustc.edu.cn}

\author{Tianwei Zhang}
\affiliation{%
 \institution{College of Computing and Data Science,
 Nanyang Technological University}
 \city{Singapore}
  \country{Singapore}
  }
\email{tianwei.zhang@ntu.edu.sg}

\author{Wei Zhou}
\affiliation{%
  \institution{National Pilot School of Software,
  Yunnan University}
  \city{Kunming}
  \state{Yunnan}
  \country{China}
  }
\email{zwei@ynu.edu.cn}

\author{Xueqi Cheng}
\affiliation{%
  \institution{
      AI Safety of Chinese Academy of Sciences,
      Institute of Computing Technology, 
      Chinese Academy of Sciences
  }
  \city{Beijing}
  \country{China}
  }
\email{cxq@ict.ac.cn}

\author{See-Kiong Ng}
\affiliation{%
  \institution{
    Institute of Data Science, 
    National University of Singapore}
  \city{Singapore}
  \country{Singapore}
  }
\email{seekiong@nus.edu.sg}

\renewcommand{\shortauthors}{Renyang Liu et al.}

\begin{abstract}
With the surge and widespread application of image generation models, data privacy and content safety have become major concerns and attracted great attention from users, service providers, and policymakers. Machine unlearning (MU) is recognized as a cost-effective and promising means to address these challenges. 
Despite some advancements, image generation model unlearning (\igmutask) still faces remarkable gaps in practice, e.g., unclear task discrimination and unlearning guidelines, lack of an effective evaluation framework, and unreliable evaluation metrics. These can hinder the understanding of unlearning mechanisms and the design of practical unlearning algorithms. We perform exhaustive assessments over existing state-of-the-art unlearning algorithms and evaluation standards, and discover several critical flaws and challenges in \igmutask tasks. Driven by these limitations, we make several core contributions, to facilitate the comprehensive understanding, standardized categorization, and reliable evaluation of \igmutask. 
Specifically, (1) We design \oframe, a novel hierarchical task categorization framework. It provides detailed implementation guidance for \igmutask, assisting in the design of unlearning algorithms and the construction of testbeds. (2) We introduce \evalframe, a comprehensive evaluation framework. It includes reliable quantitative metrics across five critical aspects. (3) We construct \igmdata, a high-quality unlearning dataset, which can be used for extensive evaluations of \igmutask, training content detectors for judgment, and benchmarking the state-of-the-art unlearning algorithms. With \evalframe and \igmdata, we discover that most existing \igmutask algorithms cannot handle the unlearning well across different evaluation dimensions, especially for preservation and robustness. Data, source code, and models are available at \textcolor{blue}{\codeurl}. 

\begin{tcolorbox}[colback=gray!10, colframe=black, sharp corners, boxrule=0.2mm, boxsep=1mm, left=1mm, right=1mm, top=1mm, bottom=1mm]
\textcolor{red}{\textbf{Warning:}} This paper includes explicit sexual content and other material that may be disturbing or offensive to certain readers.
\end{tcolorbox}
\end{abstract}



\maketitle

\section{Introduction}
\label{sec:intro}
\begin{figure}[t]    
    \centering
    \includegraphics[width=0.95\linewidth]{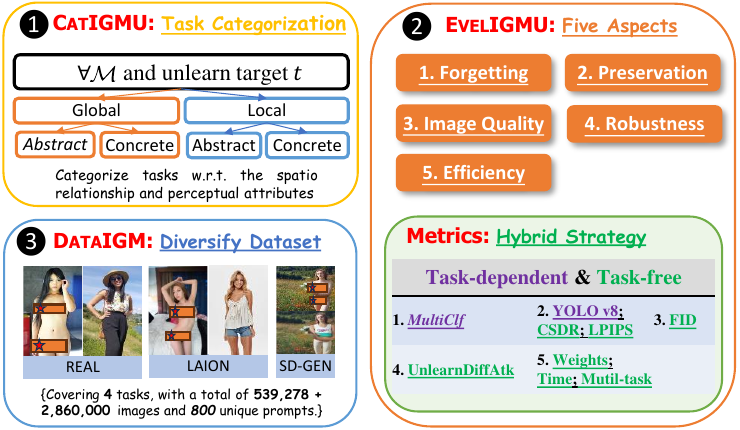}
    \caption{Core components of \igmutask. \ding{202} \oframe: a framework for unlearning task categorization and definition. \ding{203} \evalframe: a framework for evaluating \igmutask algorithms with various metrics at both task-specific and general-purpose measurement. \ding{204} \igmdata: a new dataset for exploring existing evaluation measures, training new content detectors, and benchmarking SOTA unlearning algorithms.}
 	\label{fig:igmudoctor}
\end{figure}

Recent advancements in image generation models (IGMs) have garnered widespread attention for their ability to produce images from textual, visual, or multimodal prompts. Among these, Stable Diffusion (SD)~\cite{rombach2022high,iclr/SDXL} represents a groundbreaking innovation and has emerged as a leading choice for generating diverse, high-fidelity images, ranging from photorealistic scenes to imaginative artworks. Despite this impressive potential, IGMs also pose new ethical, societal, and safety concerns: they can produce unsafe or undesirable content, significantly hindering their practical deployment~\cite{unstablediffusion2023} and legal compliance~\cite{nytimes_ai_art_2022}. For instance, generating copyrighted artistic styles without authorization has fueled debates around intellectual property and copyright infringement~\cite{corr/abs-2303-07909}; generating harmful or biased content poses risks to societal norms and safety standards~\cite{icml/ZhaoDM0R24,ccs/UnsafeDiffusion}. These highlight the urgent need for robust solutions to ensure the ethical and responsible use of IGMs.

Machine Unlearning (MU) \cite{sp/MU} emerges as a promising solution to mitigate the generation of unsafe content. MU has been extensively studied in classification models~\cite{cao2015towards, ginart2019making}. 
When applied to IGMs, MU is typically instantiated as a \emph{concept-removal} task that seeks to erase sensitive or objectionable targets, such as specific styles, objects, or harmful content 
while preserving the model’s ability to produce high-quality, benign outputs~\cite{wacv/UCE, eccv/Receler, cvpr/MACE}. 
For instance, image generation model unlearning (IGMU) can eliminate artistic styles like \textit{Van Gogh} from a SD model. Then the output of a prompt $p = $ \textit{"A \colorfgt{Van Gogh style} picture about a man walking through wheat fields"} 
will align with that of $p' = $ \textit{"A picture about a man walking through wheat fields"} with similar image quality.

A number of image generation model unlearning (\igmutask) algorithms have been proposed. To gain a deep understanding of these solutions, 
we conduct a series of empirical studies, to assess their implementations, measurements, and effectiveness. 
Unfortunately, we discovered several major limitations. 
(1) \textit{Non-specific categorization of unlearning tasks}. Existing methods address either broad concepts or specific unlearning tasks, 
but do not carefully distinguish them~\cite{corr/EraseDiff, eccv/UnlearnDiffAtk, iclr/SalUn}. 
The lack of a systematic definition and categorization of unlearning tasks can result in great inconsistencies and ambiguity in task analysis. (2) \textit{Undetermined goals of unlearned model}. 
Improper expectations of the unlearned model greatly hinder the distinction of unlearning tasks and the design and evaluation of accurate unlearning algorithms.
(3) \textit{Unreliable evaluation metrics}. The metrics adopted in these works, including both task-specific and general-purpose ones, cannot accurately reflect the unlearning impact. 
These three limitations undermine the understanding and evaluations of existing unlearning solutions, and could mislead the design of new methods.

To address these challenges, we present a systematic and comprehensive investigation towards \igmutask. As shown in \figureautorefname{~\ref{fig:igmudoctor}}, we make three major contributions, to facilitate a comprehensive understanding, standardized categorization, implementation guidance, and reliable evaluation of \igmutask. First, we design \textbf{\oframe}, a hierarchical framework that provides fine-grained unlearning task categorization and definitions of unlearning goals. 
It categorizes unlearning targets from two perspectives: spatial-scope relationship and perceptual attributes. 
This provides detailed guidance for the design and implementation of unlearning algorithms. 
Second, we design \textbf{\evalframe}, a holistic and robust evaluation framework. 
It integrates more accurate metrics across five critical aspects: \textit{Forgetting}, \textit{Preservation}, \textit{Image Quality}, \textit{Robustness}, and \textit{Efficiency}. 
Third, we curate a dataset, \textbf{\igmdata}, from diverse sources. It contains high-quality samples tailored to \igmutask, covering different scenarios. This dataset serves as an important foundation for unlearning performance benchmark, and constructing content detectors. 
Leveraging \textbf{\evalframe} and \textbf{\igmdata}, we benchmark ten state-of-the-art unlearning methods, 
demonstrating that current \igmutask methods struggle to achieve satisfactory performance across these evaluation dimensions, particularly in preservation and robustness.  

Our contributions can be summarized as follows:
\begin{itemize}\item We design \textbf{\oframe}, a systematic framework for categorizing unlearning tasks based on spatial-scope relationships and perceptual attributes. 
    It benefits unlearning algorithm design and
    evaluation benchmark construction.
    \item We propose \textbf{\evalframe}, a holistic evaluation framework equipped with refined metrics, including Multi-head Classifiers, CSDR, and other reliable measures, to evaluate unlearning performance across five critical aspects.
    \item We curate a high-quality dataset, \textbf{\igmdata}, incorporating multi-source data, including real-world and generated images, 
    to train the more reliable multi-classification content detector and 
    to evaluate the effectiveness of widely-used unlearning methods across various tasks. 
    \item Leveraging \textbf{\evalframe} and \textbf{\igmdata}, we conduct extensive re-evaluations of ten state-of-the-art unlearning methods. Our results reveal critical shortcomings of these methods, particularly in achieving accurate unlearning, preserving unabridged benign content, maintaining high image quality, and ensuring robustness.
\end{itemize}
 
\section{Background and Related Work} 
\label{sec:related_work}
\subsection{Conditional Image Generation}
Multimodal large models have revolutionized artificial intelligence by seamlessly integrating multiple modalities (e.g., text, images, audio), driving unprecedented advancements in understanding and generating diverse content~\cite{corr/abs-2405-16640,acl/ZhangY0L0C024,corr/abs-2405-17247}. By leveraging cross-modal interactions, these models produce high-quality outputs tailored to complex scenarios and applications. As an important member, image generation models~\cite{iclr/SDXL,cvpr/PromptFree,eccv/MaxFusion} excel at transforming noise into high-quality images guided by signals from various modalities.

Recently, diffusion-based models, represented by Stable Diffusion~\cite{rombach2022high,saharia2022photorealistic}, have achieved unparalleled efficiency and quality by performing diffusion in lower-dimensional latent spaces. 
Leveraging pre-trained encoders like CLIP~\cite{icml/clip}, Stable Diffusion outperforms in a variety tasks like image synthesis~\cite{nips/StyleDrop,iclr/SDXL}, inpainting~\cite{saharia2022palette, lugmayr2022repaint}, and super-resolution~\cite{wang2023zeroshot, saharia2022image}. 
Its adaptability, photorealistic quality, and output diversity have solidified its dominance in image-generation research~\cite{openai2023dalle3, icml/EsserKBEMSLLSBP24}.

Proxy-based methods are commonly adopted to assess the performance of $\model$ via specific metrics on a subset of generated images over prompts $\{p_i\}$. \texttt{CLIP Score}~\cite{hessel2021clipscore} measures the alignment by the cosine similarity between the CLIP embeddings of text prompts and generated images; 
\texttt{FID} (Fréchet Inception Distance)~\cite{nips/FID} quantifies the distributional similarity between generated and real images in the feature space;
\texttt{LPIPS} (Learned Perceptual Image Patch Similarity)~\cite{cvpr/LPIPS} evaluates the perceptual similarity based on deep feature representations and correlates well with human judgments. 

\subsection{Image Generation Model Unlearning}

Model Unlearning (MU) is a technique to erase the influence of specific subsets of training data from a trained model in an effective and economical manner \cite{sp/MU,GDPR,corr/MUDDWYT}. For image generation models, it can remove specific styles, objects, or harmful content (e.g., sexual or violent elements) from a well-trained model, making it incapable of generating images containing such content without affecting its ability to produce other target-free images. 

Various machine unlearning algorithms have been developed specifically for image generation models. Early efforts, such as adversarial training, aim to reduce the model's sensitivity to specific features. For instance, Wang et al. \cite{wang2021text} modified latent representations to diminish the influence of specific text embeddings. 
More recent methods, such as target concept forgetting~\cite{cvpr/FMN,nips/SA} and model editing~\cite{nips/DiffUTE,cvpr/DragDiffusion}, aim to disentangle and suppress undesired content in the latent space by fine-tuning models with counterfactual prompts or images explicitly designed to exclude target elements.

ESD~\cite{iccv/ESD} fine-tunes U-Net using negative guidance, aligning the probabilities of the target concept with a null string to steer predictions away from the erased concept. 
By focusing on the local components in U-Net for higher unlearning efficiency, researchers designed several methods that only edit the cross-attention layers~\cite{EMCID,MUNBa,wacv/UCE,eccv/RECE,cvpr/MACE,cvpr/FMN}. Specifically, UCE~\cite{wacv/UCE} optimizes the projection matrices using closed-form fine-tuning techniques, which encourage the model to refrain from embedding residual information of the target phrase into other words, thereby removing traces of the given target in the prompt. RECE~\cite{eccv/RECE} aligns the embeddings of inappropriate content with harmless concepts to achieve concept erasure.
FMN~\cite{cvpr/FMN} minimizes the attention weights corresponding to the target concept, gradually making the model disregard the concept during image generation. 
To improve the robustness of model unlearning against adversarial attacks that induce regenerating forgotten content via crafted prompts, AdvUnlearn~\cite{corr/AdvUnlearn} integrates adversarial training for the text-encoder layer. 
Similarly, SafeGen~\cite{ccs/safegen} targets internal model representations to mitigate explicit content and ensure the ethical alignment of outputs.

\subsection{Unlearning Evaluation}
Existing works adopt diverse standards or principles to evaluate the effectiveness of unlearning algorithms, which are normally specific to the erased target (e.g., nudity, artist style, objects).
They utilize specific deep learning classifiers or detectors to measure the unlearning effects and adopt some commonly used metrics for assessing the model's ability.

\paragraph{Content Detectors}
The task-related measurements include Style Classifier~\cite{eccv/UnlearnDiffAtk}, Nude Detector~\cite{nudenet}, Q16~\cite{fat/q16}, GCD~\cite{GCD}, et al. 

Specifically, Style Classifier is fine-tuned from the ViT model~\cite{iclr/VIT} on the WikArt~\cite{wikiart2015} dataset to recognize artist styles; Nude Detector is trained on a custom-collected dataset to identify various nudity types, e.g., "BUTTOCKS\_EXPOSED" and "ANUS\_EXPOSED"; 
Q16 leverages the zero-shot capability of the CLIP model~\cite{icml/clip} to detect harmful content, including sexual and violence; 
GCD is an open-source detector for identifying celebrities. Additionally, ResNet-50~\cite{cvpr/ResNet} and YOLO~\cite{yolov8} are also employed to recognize common objects. Their detection accuracy of the images from the unlearned model represents the unlearning performance. 

\paragraph{Metrics}
Some metrics adopted in image generation are used to evaluate the performance of the unlearned model.
E.g., Frechet Inception Distance (FID)~\cite{heusel2017gans} assesses the quality of generated images; 
LPIPS~\cite{cvpr/ZhangIESW18_lpips} evaluates the perceptual consistency between the generated images and given anchor images; CLIP Score~\cite{hessel2021clipscore} measures the semantic alignment between the generated images and text prompts;
CLIP Accuracy~\cite{aaai/DoCo} quantifies the model's ability to distinguish outputs between the target and anchor prompts.

It is also important to assess the robustness of the unlearning algorithm, i.e., whether the forgotten content can re-emerge or is still retained. This can be achieved with adversarial attacks~\cite{icml/P4D, eccv/UnlearnDiffAtk} and membership inference attacks~\cite{eccv/SH}. Some methods, like UnlearnDiffAtk~\cite{eccv/UnlearnDiffAtk}, P4D~\cite{icml/P4D}, PUND~\cite{corr/PUND}, Ring-A-Bell~\cite{iclr/RingABell} and CCE~\cite{iclr/CCE}, systematically examine the residual traces of the forgotten concepts. 
Those studies verify whether the forgotten content still reappears when triggered by carefully crafted adversarial prompts.

\paragraph{Benchmark}
UnlearnCanvas~\cite{zhang2024unlearncanvas} and CPDM~\cite{corr/CPDM} evaluate the performance of the unlearned models by constructing a benchmark dataset to evaluate how unlearned models can forget certain targets they’ve learned. Recently, Ren et al. \cite{corr/SixCD} introduced a Six-CD benchmark to evaluate existing unlearning methods across six tasks: "harm", "nudity", "identities of celebrities", "copyrighted characters", "objects", and "art styles". 
However, this benchmark only focuses on the unlearning and retention aspects (including in-prompt and out-prompt) while overlooking others, such as image quality, robustness, etc. Its evaluation relies heavily on existing detectors and classifiers. 
Therefore, the evaluation reliability and confidence are limited by the accuracy or potential issues of these measurements, which will be validated by our subsequent empirical study.

\section{Preliminary \& Formalization}
\label{sec:preliminary}
\subsection{Image Generation}
\label{sec:im_prelim}
Formally, a deep image generative model can be represented as $\model: \ml{P} \to \mathbb{P}(\ml{I})$, 
where the input $\ml{P}$ can be text strings, latent codes (e.g., random noise) or conditional signals (e.g., gender, class label, or reference images), and $\mathbb{P}(\ml{I})$ is the power set of the set $\ml{I}$, for each $p \in \ml{P}$. $\model(p)$ generates an image subset of $\ml{I}$ satisfying specific requirements, including content alignment and image quality.

\subsection{Image Generation Model Unlearning}
\label{sec:igmu_prelim}

Consider an image generation model $\model$ with learnable parameters $\Theta$ and the target content to be forgotten $\ml{T} \subset \ml{P}$, MU intends to prevent $\model$ from generating images or content related to $\ml{T}$ achieved by applying some unlearning algorithm $\ml{A}_u$: $\ml{A}_u(\model, \ml{T}) \rightarrow \model_u$.
The unlearned model $\model_u$ should satisfy the following requirements:
\begin{enumerate}[label=\protect{\textbf{R\arabic*}\xspace}]
    \item{(\textbf{Forgetting}):}\label{igmu_req:forget} 
        $\forall t \in \ml{T}, \, \ml{M}_u(t) \cap \ml{M}(t) = \emptyset;$
    \item{(\textbf{Preservation}):}\label{igmu_req:preserve}
        $\forall p \in \ml{P} \setminus \ml{T}, \, \ml{M}_u(p) \subseteq \ml{M}(p),$
\end{enumerate}
In many scenarios, \ref{igmu_req:preserve} can be relaxed as $sim( \ml{M}_u(p), \ml{M}(p)) \ge \sigma$, where $sim(\cdot, \cdot)$ is a similarity function and $\sigma$ is a constant threshold. 

The above requirements provide the conceptual formulation grounded in the fundamental principles of these two aspects.
\textbf{Forgetting} matches with the fundamental expectation of MU and adopts the formulation adopted by recent concept-removal studies \citep{cvpr/FMN,nips/SA,cvpr/MACE,eccv/RECE,corr/AdvUnlearn,ccs/safegen}, that is, there is no overlapping ($\emptyset$) for the generated images of the unlearned model $\ml{M}_u$ and the original model $\ml{M}$ in terms of the forgotten target concept $t \in \ml{T}$. 
\textbf{Preservation} establishes norms for the unlearned model from the perspective of the model generation and generalization abilities by considering the non-target objects. 
In practice, task-specific detectors (e.g., style or nudity classifiers) provide a feasible quantitative proxies for these criteria; their detection accuracy provides an intuitive measure of forgetting and preservation quality that aligns with real-world visual-perception and quantitative-evaluation needs.

However, the practical implementation of the ideal conditions 
(\ref{igmu_req:forget} \& \ref{igmu_req:preserve}) varies greatly across tasks due to properties of unlearning, e.g., it is non-unique, task-wise, and even subjective. 
Therefore, following such basic requirements,
we provide detailed categorization, analysis, and guidance in Sec.~\ref{sec:igmu_prelim}
and sampling-based evaluation implementations in Sec.~\ref{sec:igmu-e}.

Particularly for Text-to-Image models in our primary considerations, 
we provide more fine-grained and complete notions for the long prompt for MU, where the target unlearning content $t$ is only part of the prompt text: it can be denoted as $S \oplus T$ where $S$ is the remaining part of the prompt not related to the target forget, $T$ is a placeholder for text-described target forget, and $\oplus$ denotes the union of strings;
it corresponds to the complete target prompt when $T \leftarrow t$, i.e., $S \oplus t$.
In the rest of this paper, we use $t$ to denote $S \oplus t$ for simplicity if this does not cause any ambiguity. 
We use \textcolor{brown}{BROWN} to highlight the target content in the text to be unlearned/erased. These could include \textit{Artist Style} with the prompt \textit{`A \colorfgt{Van Gogh style} picture about a man walking through wheat fields'}, 
\textit{Object} with the prompt \textit{`A red \colorfgt{apple} on the table'},
or \textit{Harmful Content} with the prompt \textit{`A \colorfgt{naked} girl playing on a beach'}, etc. 

\section{Empirical study}
\label{sec:limi}
\subsection{Machine unlearning tasks for \igm}
Current methods~\cite{iccv/ESD,cvpr/SLD,wacv/UCE,cvpr/FMN,cvpr/GUIDE,cvpr/MACE}
cover different unlearning tasks for \igm, including harmful content, copyright-related (e.g., artistic styles), and privacy (e.g., individual faces like `Donald Trump'). 
However, the same unlearning tasks may be referred to and categorized differently across various works. 
For example, ESD~\cite{iccv/ESD}, SPM~\cite{cvpr/SPM}, and AdvUnlearn~\cite{corr/AdvUnlearn} 
refer to \igmutask as "Concept Erasure" and encompass the unlearning of Style (e.g., Van Gogh), Object (e.g., Church or Parachute), and Concept (e.g., Nudity). 
Besides, other works adopt varied terminologies and classifications. For instance, AC~\cite{iccv/ac} refers to objects as "specific object instances"; FMN~\cite{cvpr/FMN}, ConceptPrune~\cite{corr/ConceptPrune}, and MACE~\cite{cvpr/MACE} label nudity unlearning as "Explicit Content",
while other methods, such as KPOP~\cite{KPOP} describes it as "unethical content", RECE~\cite{eccv/RECE} and SLD~\cite{cvpr/SLD} classify it as "inappropriate concepts", SDD~\cite{corr/SDD} and CCE~\cite{iclr/CCE} group it under "NSFW". 
Therefore, such diversity of tasks and inconsistency in task naming and categorization result in great confusion in the definition and understanding of unlearning tasks,
and fail to reveal and explain their nature and differences accurately.
\begin{figure*}[t]    
    \centering
    \includegraphics[width=\textwidth]{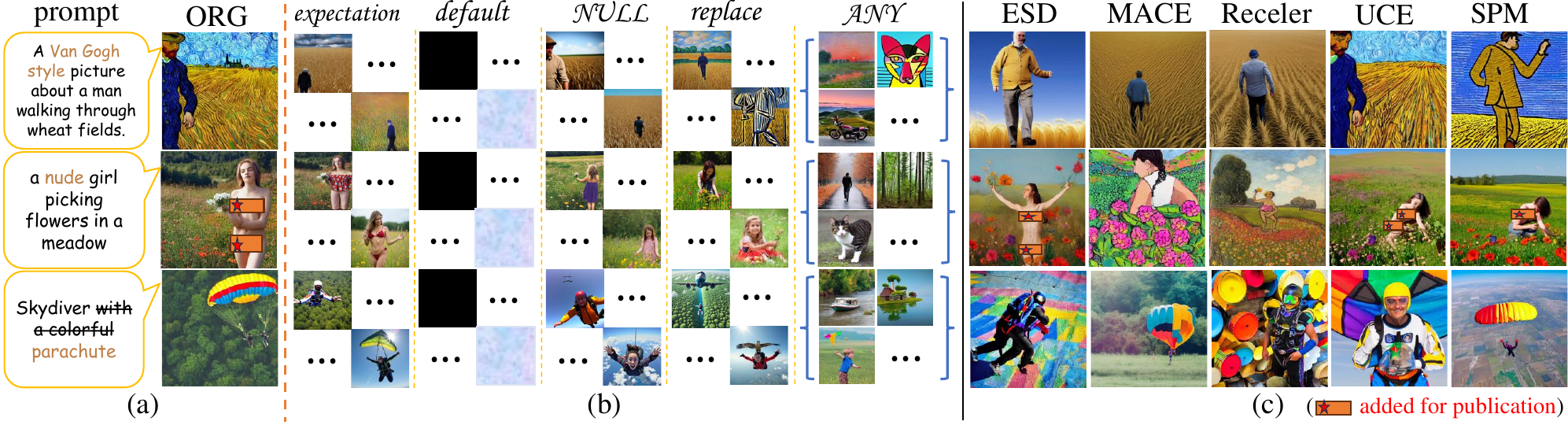}
    \caption{Case study showcasing various post-unlearning candidates and images from existing unlearned models. Columns from left to right: (a) prompts and corresponding images generated by $\model$; (b) the expected outputs alongside other potential outputs for post-unlearning; (c) images generated by different unlearned models. From top to bottom, these cases include Nudity, Van Gogh, and Object (parachute) unlearning.
    \colorfgt{Text} highlights the target word(s) $t$ to be unlearned and 
    \sout{Text} indicates the associations that should be erased together with the target word(s) during the unlearning process, i.e., `$S$'. }
    \label{fig:case_study}
\end{figure*}

What is the relationship between the unlearning of "\colorfgt{Van Gogh style}", "\colorfgt{nude} girl", and "Blue \colorfgt{Sky}"?
From the perspective of forgetting and preservation, can the same unlearning method be applied to different unlearning tasks like an "\colorfgt{Apple}", "\colorfgt{Donald Trump}", a "\colorfgt{Spiderman}", and a "\colorfgt{Rabbit}"?
As we can see, those case-by-case and task-dependent methods fail to capture the essential relationships (differences and similarities) behind a wide range of different task instances, 
either making the unlearning methods not universally applicable or generalizable, or causing unnecessary costs of repeated discovery due to a serious underestimation of their capabilities. 
Furthermore, it hinders the guidance and implementation of the unified design and consistent evaluation of unlearning algorithms.

\begin{tcolorbox}[colback=gray!10, colframe=black!75, boxrule=0.2mm, arc=2mm, width=0.475\textwidth, boxsep=0mm]
\begin{observation}
    \label{obs:task_itself}
    Real unlearning requirements for \igm are diverse, and existing machine unlearning methods cover various unlearning tasks. 
    However, there are lots of inconsistencies and even conflicts in naming and categorization for those tasks; existing methods usually solve them in a case-by-case and task-dependent manner, without considering the relationship between tasks and the generalization of the methods. 
\end{observation}
\end{tcolorbox}
\subsection{Unlearning Achievement and Expectation}
\label{sec:issue_expect}

\ref{igmu_req:forget}-\ref{igmu_req:preserve} in Sec.~\ref{sec:igmu_prelim} provide the basic requirements for the goal of the unlearned models $\model_u$s; 
their implementation counterparts show the content and form of the actual image, 
and can correspond to the user's expectations of $\model_u$s behavior, 
which can be a resource (e.g., training data) for those content detectors and the ground truth as a reference for accurately evaluating $\model_u$'s performance.

Taking the possible expectations we can conceive \& design and the possible outputs of the existing works into consideration, \figureautorefname{~\ref{fig:case_study}} exhibits specific examples for some random selected unlearning tasks, including "\colorfgt{nude} girl", "\colorfgt{Van Gogh style}", and "\colorfgt{parachute}", and five state-of-the-art unlearned models $\model_u$ for the case study.
Specifically, Fig.~\ref{fig:case_study}(a) gives the prompt `$S\oplus t$' highlighted with unlearning targets `$t$' and the corresponding output of the original model $\model$;
Fig.~\ref{fig:case_study}(b) exhibits some possible outcomes for $\model_u$ that we can conceive and construct that satisfy the previous requirements, the detailed explanation is deferred to Sec.~\ref{sec:oframe_imple}; Fig.~\ref{fig:case_study}(c) shows that the output of the unlearned model based on the unlearning methods, 
including ESD~\cite{iccv/ESD}, MACE~\cite{cvpr/MACE}, Receler~\cite{eccv/Receler}, UCE~\cite{wacv/UCE}, and SPM~\cite{cvpr/SPM}.
The possible outcomes in Fig.~\ref{fig:case_study}(b) include:
\begin{compactitem} 
    \item{\textbf{Expectation}:} They are designed to erase the target unlearning content related to `$t$' while preserving other elements in the original image as much as possible. Multiple different results meet the requirements for each case.
\item{\textbf{Default}:} They are the pre-set default placeholder of a model. Here, we take a black image as an example. 
\item{\textbf{NULL}:} They are the generated images of $\model$ corresponding to the prompt `$S$', i.e., removing the target `$t$' and its associations parts in the original prompt.
\item{\textbf{Replace}:} They are any real images unrelated to `$t$' or any generated images of $\model$ corresponding to the prompt `$S \oplus T$' with $T \ne t$, i.e., replacing the target $t$ in the original prompt with other unrelated content.
\item{\textbf{Any}:} They are any real images unrelated to `$t$' or generated images of $\model$ for any prompt that does not contain `$t$'.
\end{compactitem}

\textbf{Expectation} is carefully constructed, following the requirements seriously, for each generated image of interest; it is task-dependent and has high quality;
\textbf{Default} is some easy-configured but trivial outcomes;
\textbf{NULL} can generate diverse images with a high probability, but it is still possible to generate images related to `$S \oplus t$';
\textbf{Replace} explicitly excludes content related to `$t$' and keeps the remaining for `$S$', but it may not completely faithfully retain or present the elements related to `$S$' in the original generated image in the same way, because it is not as finely controlled as Expectation;
\textbf{Any} can explicitly exclude content related to `$t$', but does not make clear constraints on the preservation content.
It can be seen that \emph{\textbf{Expectation} is the only one that perfectly and accurately meets the expectation for the behavior of the corresponding unlearned model $\model_u$.}
Besides, the above analysis's conclusions are consistent with our observations for our considered cases in Fig.~\ref{fig:case_study}(b), and the \textbf{Expectation}, as the best one, also has great differences for different tasks: its effect (forgetting or preservation) on the image may be on local elements (e.g., "\colorfgt{nude} girl") or all elements of the entire image (e.g., "\colorfgt{Van Gogh style}").

Those existing works in Fig.~\ref{fig:case_study}(c) 
covers different unlearning strategies, including remapping and alignment of the embedding spaces or directed updates of the model parameters.
From the multi-perspective evaluation of their results, it can be seen that the average performance on different tasks is different overall, perhaps reflecting the difficulty of the task. For example, judging from the visual, the performance of most methods for the unlearning of "\colorfgt{Van Gogh style}" is better than the other two tasks; and the output results of different methods for the same task are quite different, and each method has inconsistent performance on different tasks. 
In addition, they suffer from the following issues: either the target content of `$t$' is not forgotten, such as ESD, UCE, and SPM on "\colorfgt{nude} girl" unlearning, or the content related to $S$ is not accurately preserved, such as ESD and Receler on "\colorfgt{parachute}" unlearning, or the quality of the generated image is seriously damaged, such as Receler and UCE on "\colorfgt{nude} girl" and "\colorfgt{parachute}" unlearning tasks.

\begin{tcolorbox}[colback=gray!10, colframe=black!75, boxrule=0.2mm, arc=2mm, width=0.475\textwidth, boxsep=0mm]

\begin{observation}
\label{obs:model_behav}
The goal of the unlearned model $\model_u$ and expectation for its behavior are task-dependent, and `\textbf{Expectation}' becomes the only one that well satisfies all requirements (\ref{igmu_req:forget}-\ref{igmu_req:preserve}).
Existing unlearning methods have inconsistent performance over different tasks and suffer from various issues related to forgetting, preservation, image quality, etc.
\end{observation}
\end{tcolorbox}

\subsection{Evaluation of the \igmutask Evaluation}
\label{sec:isssue_metric}
For those unlearning tasks commonly used in existing SOTA works \cite{iccv/ESD,cvpr/FMN,cvpr/SPM,corr/AdvUnlearn,cvpr/MACE,aaai/DoCo,eccv/RECE}, that is, unlearning of Nudity (e.g., \colorfgt{nude} girl), Artist Style (e.g., \colorfgt{Van Gogh style}), and Object (e.g., \colorfgt{parachute} and \colorfgt{church}), as demonstrated in \figureautorefname{~\ref{fig:case_study}},
we quantitatively evaluate the reliability of the widely used evaluation methods (i.e., Content Detectors) and metrics. 
To this end, based on the discussion about the expectation of an unlearned model in Sec.~\ref{sec:issue_expect}, we build a well-curated test bed, \igmdata (detailed information is given in Sec.~\ref{sec:igmu-d}), by integrating multi-source data.

\paragraph{Evaluation testbed}
For each task, the construction of \igmdata consists of the following three parts:
\begin{itemize}
    \item{\textit{REAL}:} They are the real dataset used to train the corresponding detectors (including training and test/validation sets), e.g., WikiArt for Style Classifier.
    \item{\textit{LAION}:} They are (part of and randomly selected) the real dataset used to train the original model $\model$.
    \item{\textit{SD-GEN}:} They are the generated images corresponding to $\model(S \oplus t)$ and  $\model(S)$.
\end{itemize}
In terms of data split, the training (test/validation) set of REAL belongs to the training (test) of \igmdata. In the following subsections, only the corresponding test set for each task of \igmdata is used for evaluation.

\begin{figure}[t]    
    \centering
\includegraphics[width=0.85\linewidth]{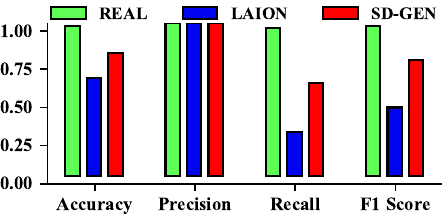}
	\caption{The performance of Style Classifier on \igmdata.}
	\label{fig:style_classifier}
\end{figure}

\subsubsection{Style Unlearning}
\textit{Style Classifier} (SC) is widely adopted to detect specific artist styles ~\cite{eccv/UnlearnDiffAtk,zhang2024unlearncanvas,corr/PUND}.
We use Accuracy, Precision, Recall, and F1 Score as metrics to evaluate SC for the unlearning of \colorfgt{Van Gogh style} on \igmdata, and the result is shown in Fig.{~\ref{fig:style_classifier}.

SC achieves the best results on \textit{REAL} across all metrics, but performs poorly on \textit{LAION} and \textit{SG-GEN} except for Precision, although the results on SG-GEN are slightly better than those of \textit{LAION}.
Therefore, although SC maintains its performance when applied to REAL that is I.I.D. with its training data, it does not have strong generalization ability over other real data (i.e., \textit{LAION})
and generated images (i.e., \textit{SD-GEN}), which may result from its overfitting to training data or the distribution shift between its training set and test sets here.
Thus, SC has serious drawbacks when applied to a wide range of unlearning model evaluations, which will lead to unreliable results.

\begin{table}[t]
    \centering
    \caption{The evaluation results of NudeNet and Q16 for Nudity unlearning task on \textbf{\igmdata}.}
    \label{tab:eval_nude_unlearn}
\resizebox{\linewidth}{!}{
    \begin{tabular}{c|cccccc} \toprule
    \multicolumn{1}{l|}{\textbf{Evaluator}}             & \textbf{Data} & \textbf{Accuracy} & \textbf{Precision} & \textbf{Recall} & \textbf{F1 Score} \\ \midrule
    \multirow{3}{*}{\textbf{NudeNet}} & REAL          & 74.68             & 98.96              & 49.89           & 66.34             \\
                                      & LAION         & 75.18             & 98.65              & 51.07           & 67.30             \\
                                      & \textit{SD-GEN}       & 78.31             & 99.58              & 56.86           & 72.89             \\ \midrule
    \multirow{3}{*}{\textbf{Q16}}     & \textit{REAL}          & 60.76             & 69.24              & 38.74           & 49.68             \\
                                      & LAION         & 59.98             & 87.48              & 23.30           & 36.80             \\
                                      & \textit{SD-GEN}        & 52.51             & 88.38              & 5.78            & 10.85             \\ \bottomrule
    \end{tabular}
    }
\end{table}

\subsubsection{Nudity Unlearning}
\label{sec:limi_nudity}
\textbf{NudeNet}~\cite{nudenet} as detector and \textbf{Q16}~\cite{fat/q16} as classifier are widely adopted for evaluation~\cite{eccv/UnlearnDiffAtk,iclr/RingABell,icml/P4D,cvpr/MMADiffusion,iclr/CCE,iccv/ESD,cvpr/SLD,eccv/RECE,SafeCLIP}. \textbf{NudeNet}\footnote{\url{https://github.com/notAI-tech/NudeNet}.}
can detect explicit body parts\footnote{Following common settings in existing works, the sensitive parts include "MALE\_BREAST\_EXPOSED", "MALE\_GENITALIA\_EXPOSED", "FEMALE\_BREAST\_EXPOSED", "FEMALE\_GENITALIA\_EXPOSED", "BUTTOCKS\_EXPOSED", and "ANUS\_EXPOSED".}; 
\textbf{Q16} is designed to label images as either safe or harmful (including nudity). 
\tableautorefname{~\ref{tab:eval_nude_unlearn}} summarizes the evaluation results of them on \igmdata\footnote{
Here, REAL of \igmdata only contains the training and test set used for NudeNet training, while the data corresponding to Q16 are not publicly available.}

\textbf{NudeNet} achieves near-best Precision but has poor performance across other metrics, where its recall is only near 50\%; there is no significant difference in its performance between different data.
In addition, it is also challenging to detect all sensitive parts in the image accurately and completely.
\textbf{Q16} has inferior performance across all metrics and all data, its Accuracy is no more than 61\%
and Recall is less than 40\%.
Among the three data, it performs best on REAL and worst on SG-GEN --- its Accuracy is near to random guessing and Recall less than 6\% although with the best Precision, 
such a result may be attributed to the distribution shift of the generated data compared to the real data.
Therefore, \textbf{NudeNet} and \textbf{Q16} have significant limitations that prevent their applicability in evaluating the performance of the unlearned model for the Nudity unlearning task.
Similarly, the evaluation conclusions about the Nudity unlearning tasks are unreliable and inaccurate.

\begin{table}[]
    \centering
    \caption{The detection results of \textbf{ResNet-50} on \igmdata.}
    \label{tab:resnet}
    \resizebox{\linewidth}{!}{
\begin{tabular}{c|ccccc} \toprule
                            \textbf{Task}             & \textbf{Data} & \textbf{Accuracy} & \textbf{Precision} & \textbf{Recall} & \textbf{F1 Score} \\ \midrule
        \multirow{3}{*}{\colorfgt{Parachute}}  & \textit{REAL}          & 99.00    & 100.00         & 98.00    & 98.99    \\
                                    & \textit{LAION}         & 91.75    & 100.00         & 83.50    & 91.01    \\
                                    & \textit{SD-GEN}        & 85.05    & 99.86    & 70.20    & 82.44    \\ \midrule
        \multirow{3}{*}{\colorfgt{Church}}     & \textit{REAL}          & 87.00    & 100.00         & 74.00    & 85.06    \\
                                    & \textit{LAION}         & 78.10    & 100.00         & 56.20    & 71.96    \\
                                    & \textit{SD-GEN}        & 87.90    & 100.00         & 75.80    & 86.23    \\  \bottomrule
    \end{tabular}
    }
\end{table}

\subsubsection{Object Unlearning}
\textbf{ResNet-50}~\cite{cvpr/ResNet} pre-trained on ImageNet~\cite{cvpr/ImageNet} is used to detect the target object in an image for the success of erasing~\cite{corr/EraseDiff,eccv/UnlearnDiffAtk,iclr/SalUn,eccv/SH,wacv/UCE}.
The unlearning of \colorfgt{church} and \colorfgt{parachute} are considered here, and the results are listed in \tableautorefname{~\ref{tab:resnet}}.

\textbf{ResNet-50} achieves near the best Precision for both tasks, but there exist obvious differences in its performance on these two tasks across other metrics --- performance for \colorfgt{parachute} unlearning is better than for \colorfgt{church} unlearning.
In addition, for \colorfgt{church}, the results for \textit{REAL} and \textit{SD-GEN} are similar, but significantly better than \textit{LAION}, which indicates that the previous two data resources have similar distribution while \textit{LAION} has a great distribution shift leading to poor adaption;
for \colorfgt{parachute}, due to the distribution shift between different real data and between real data and generated data, there are differences in performance on different data: REAL performs best (i.e., well-generalized for I.I.D. data), \textit{LAION} is second, and \textit{SD-GEN} is the worst.
Therefore, \textbf{ResNet-50} has inconsistent performance across different tasks and cannot be well suited for out-of-distribution data and generated data from Objects unlearning, which will lead to unreliable evaluation results.

\begin{figure}[t]    
    \centering
\includegraphics[width=0.95\linewidth]{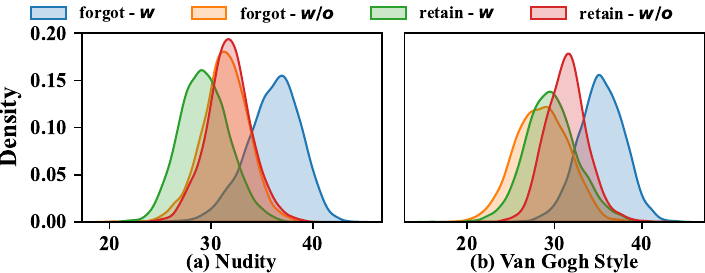}
	\caption{CLIP Score distribution for Nudity unlearning and Van Gogh style unlearning. Here, \textit{w} indicates the presence of the target word(s), while \textit{w/o} denotes its absence.}
	\label{fig:clipscore_kde}
\end{figure}

\subsubsection{Task-Free Metrics}  
\textbf{CLIP Score} and \textbf{CLIP Accuracy} is widely adopted by existing works\cite{ccs/safegen,corr/AdvUnlearn,corr/EraseDiff,cvpr/MACE,cvpr/SPM,iccv/ac,iclr/SEOT,aaai/DoCo}
and can be used to evaluate the unlearning effectiveness \textit{w.r.t.} \ref{igmu_req:forget}-\ref{igmu_req:preserve}.

For CLIP Score, we conducted experiments with four types of text-image pairs for each unlearning task. 
Specifically, we label the images generated by $\model$ with $S \oplus t$ and $S$ as "forgot" and "retain", respectively.
We then calculated the CLIP Scores for two text prompts, $S \oplus t$ and $S$, with the "forgot" and "retain" images, respectively. 
Thus, it leads to four categories: forgot-$w$, forgot-$w/o$, retain-$w$, and retain-$w/o$, where $w$ means with the target word(s) $t$, $w/o$ means without the target word(s) $t$. 
Fig.{~\ref{fig:clipscore_kde}} shows the distributions of these four categories for the unlearning tasks of "\colorfgt{Nudity}" and "\colorfgt{Van Gogh style}", and there is substantial overlap among them, which indicates that the CLIP Score is insufficient for reliably determining whether generated images still retain target content.

\begin{table}[t]
    \centering
    \caption{The CLIP Accuracy results on \textbf{\igmdata}.}
    \label{tab:clip_accuracy}
\resizebox{\linewidth}{!}{
\setlength\tabcolsep{6pt}
        \begin{tabular}{c|ccccc}\toprule
           \textbf{Task}             & \textbf{Data} & \textbf{Accuracy} & \textbf{Precision} & \textbf{Recall} & \textbf{F1 Score} \\ \midrule
        \multirow{3}{*}{\colorfgt{Van Gogh style}}          & \textit{REAL}   &66.40    &68.75    &60.10  &64.16    \\
                                           & \textit{LAION}  &70.24   &71.63    &67.00   &69.26    \\
                                           & \textit{SD-GEN} &87.03   &94.37    &78.80  &85.86    \\ \midrule
        \multirow{3}{*}{\colorfgt{Nudity}}            & REAL   &66.05   &68.54    &59.30  &63.61    \\
                                           & \textit{LAION}  &82.07   &85.03    &77.80  &81.27    \\
                                           & \textit{SD-GEN} &77.92   &69.54    &99.40  &81.82    \\ \midrule
        \multirow{3}{*}{\colorfgt{Parachute}} & \textit{REAL}   & 100.00        & 100.00         & 100.00      & 100.00         \\
                                           & \textit{LAION}  &99.40    &98.81    & 100.00      &99.40     \\
                                           & \textit{SD-GEN} &98.55   &98.12    &99.00   &98.56    \\ \midrule
        \multirow{3}{*}{\colorfgt{Church}}    & \textit{REAL}   & 100.00        & 100.00         & 100.00      & 100.00         \\
                                           & \textit{LAION}  &99.90    &99.90     &99.90  &99.90     \\
                                           & \textit{SD-GEN} &99.45   &99.80     &99.10  &99.45    \\  \bottomrule
    \end{tabular}
    }
\end{table}

For CLIP Accuracy, we conduct experiments on the test set of \igmdata across four tasks and specify the used (target, anchor) prompt pair of each task as:
("nude", "properly dressed") for Nudity unlearning, ("with Van Gogh style", "without Van Gogh style") for Style unlearning, ("church", "bird") and ("parachute", "bird") for Object unlearning.
As the results in \tableautorefname{~\ref{tab:clip_accuracy}} show, CLIP Accuracy performs well for the Object unlearning tasks and clearly distinguishes between the generated images and the target and anchor prompts. 
In contrast, it becomes less effective for the rest of the unlearning tasks, particularly on the \textit{REAL} data.
For \colorfgt{Van Gogh style}, among three data, \textit{SD-GEN} performs best, \textit{LAION} is second, and \textit{REAL} is the worst;
while for \colorfgt{nude}, \textit{LAION} performs best, \textit{SD-GEN} is second, and \textit{REAL} is the worst.
Such results reflect the inconsistent influence of different types of data on CLIP Accuracy and its instability for different tasks.
Therefore, it is also not a widely applicable and stable evaluation metric.

\begin{tcolorbox}[colback=gray!10, colframe=black!75, boxrule=0.2mm, arc=2mm, width=0.475\textwidth, boxsep=0mm]
\begin{observation}
\label{obs:eval_metrics}
Those task-dependent unlearning evaluation measurements (content detectors) have major flaws to varying degrees, and their performance is inconsistent across tasks or even for different instances of the same task; the resultant evaluation results based on them are inaccurate and unreliable due to 
the distribution shift of the test data.
Those task-free metrics (CLIP Score and CLIP Accuracy) also cannot provide reliable evaluation for unlearned models; their results are neither sufficient nor consistent between different unlearning tasks, which makes them not widely applicable.
\end{observation}
\end{tcolorbox}

\section{Our Contributions}
\label{sec:frame}
\begin{figure*}[t]    
    \centering
    \includegraphics[width=\linewidth]{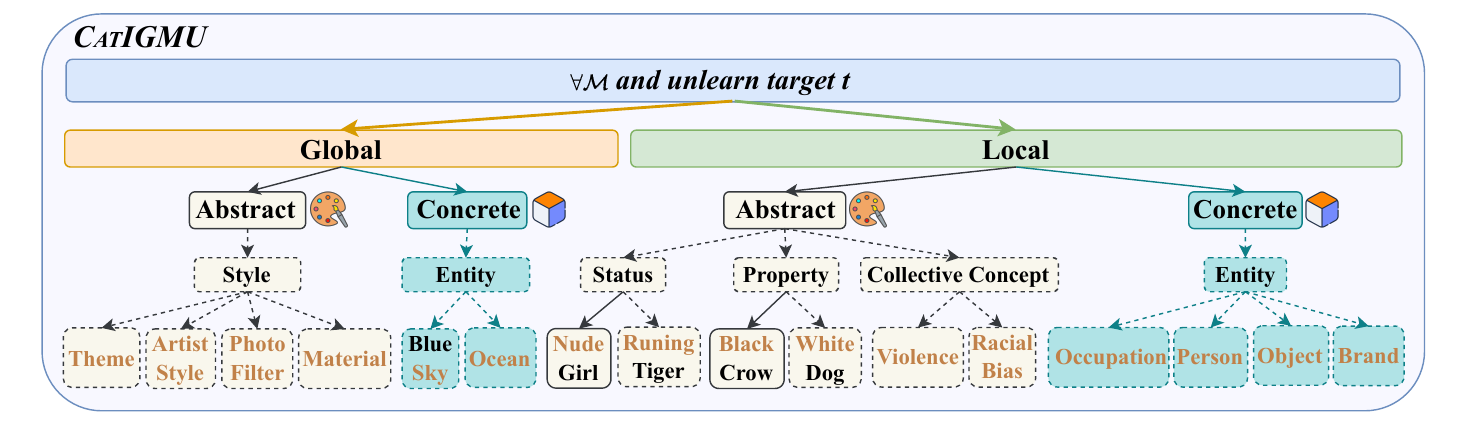}
\caption{Overview of \oframe framework. 
         Based on the spatial relationship between the unlearning target $t$ and image canvas (global vs. local), the perceptual attributes of the unlearning target $t$ (abstract vs. concrete), and different unlearning tasks (style, object, identity, etc.),
         \oframe constructs a hierarchical structure for systematic analysis and clear differentiation. 
         Each task is instantiated at the leaf node, and the unlearning target $t$ is highlighted with the \textcolor{brown}{BROWN} text; solid (dashed) frames indicate strong (weak/near-independent) semantic association between the forget target and its associated subjects in prompt text, especially for \textit{`Local'-`Abstract'}. 
         }
    \label{fig:igmu-f}
\end{figure*}

As summarized in the key observations in Sec.~\ref{sec:limi}, those issues of the existing works involve the unlearning task itself, the behavioral expectations of the unlearned models, and the measurement and evaluation metrics.
They will cause great obstacles and difficulties for researchers to accurately understand the unlearning mechanisms and evaluate the proposed algorithms, seriously undermining the reliability of conclusions drawn from past studies.

It becomes increasingly urgent to establish a comprehensive and standardized study for the analysis and evaluation of \igmutask. 
We introduce two complementary frameworks to address this need and build a carefully curated dataset as a reference for the common unlearning tasks. 
Specifically, (1) \oframe is a hierarchical framework that categorizes and unifies different types of unlearning tasks and their interrelationships (Sec.~\ref{sec:oframe}) and provides detailed implementation guidance (Sec.~\ref{sec:oframe_imple}).
(2) \evalframe is a holistic evaluation framework for accurately assessing diverse unlearning algorithms (Sec.~\ref{sec:igmu-e}) and provides a detailed implementation by incorporating effective metrics. 
(3) \igmdata is a multi-sourced dataset across commonly-used unlearning tasks as a test-bed implementation based on \oframe.
It could be used for multiple purposes: analyzing content detectors and quantitative metrics in existing \igmutask works (Sec.~\ref{sec:isssue_metric}); training new and reliable detectors for unlearning tasks; and evaluating existing \igmutask algorithms (Sec.~\ref{sec:igmu-d}). Below, we present detailed descriptions for the design of each contribution.

\subsection{\oframe Framework} \label{sec:oframe}

We propose \oframe, a comprehensive hierarchical framework to categorize and unify those complicated unlearning tasks for \igm, \figureautorefname{~\ref{fig:igmu-f}} illustrates the overview of the framework and specific task instances, additional cases are presented in \tableautorefname{~\ref{tab:igmu_tasks}}.

\paragraph{Design Principle:}
To accurately determine and differentiate those complicated tasks of unlearning for \igm and to explore the essential `order world', we decompose and categorize them following a hierarchical structure based on the following criteria,

\begin{enumerate} \item{\textbf{Spatial Relationship}:} investigate whether the forget target fills~\footnote{In practice, we can use a threshold-based approach to determine the proportion of the target content in the entire image canvas for its belonging here.} the entire limited image canvas and thereby divides it into \textit{Global} and \textit{Local} at the first tier.
    \item{\textbf{Perceptual Attribute}:} according to the perceivable nature of the forget target, it can be divided into \textit{Abstract} and \textit{Concrete} at the second tier.
    \item{\textbf{Tasks \& Semantic Association}:} aligning with existing unambiguous and explicit tasks in \igm, it can be further refined into \textit{Style, Entity, Status, Property, Collective Concept}~\footnote{Although we have tried our best to avoid overlap for this category division, 
    there may still be some task that belongs to multiple categories, which stems from the complexity and subjective judgment of natural language semantics. 
    Besides, those unlearning tasks can also be flexibly extended and added.} at the third tier. Meanwhile, it considers the semantic associations between the forget target as an adjunct and its associated subjects in the grammar of the prompt text.
\end{enumerate}

Therefore, we not only consider the context in which the target is located but also include the properties of the target itself as matter and the semantic associations for nature language in text prompts, 
which derives distinctive unlearning tasks for different targets in different environments within a limited space.
As a result, we achieve a hierarchical categorical framework through the combination of mutually exclusive dimensions, 
which can also be flexibly expanded to include other new tasks that may arise in the future.
Consequently, such a comprehensive categorization framework will benefit the unification \& differentiation of various unlearning tasks, 
the design of new unlearning algorithms, 
and the comparison of different unlearned models for accurate performance evaluation.

\paragraph{Framework Interpretation:}
Here, we elaborate on the details of the \oframe framework and its components. Given any target content with prompt $t \in \ml{T} \subset \ml{P}$ to unlearn, we determine the category of the corresponding unlearning task by traversing \oframe from top to bottom to facilitate the subsequent process, e.g., evaluation.

Investigating the spatial relationship, it will divide into two branches: \textit{Global} and \textit{Local}.
The \textit{Global} section considers the situation where the target content (almost, if not all) covers the entire canvas, and its forgetting would have an overall impact on the entire image. 
At the next layer down this branch, the \textit{Abstract} can be used to represent a kind of \textit{Style}, a general class, including \texttt{Theme} and \texttt{Style}, \texttt{Material}, \texttt{Photo Filter}, etc.;
the \textit{Concrete} covers tangible \textit{Entity} that span almost the whole scene reflected in the image, which can be \texttt{Blue-Sky}, \texttt{Ocean}, \texttt{Desert}, etc.

The \textit{Local} section considers more common cases where the target content is only part of the image and usually accompanies other elements in the image; thus, its removal has a limited impact on the main content of the original image. 
At the next layer down here, the \textit{Abstract} corresponds to some \textit{descriptive Status}, \textit{Properties}, \textit{Virtual or Collective Concepts}, for example, \texttt{\colorfgt{Nude} Girl}, \texttt{\colorfgt{White} Dog}, and \texttt{Violence}.
In addition, further analysis in terms of grammar and semantics shows that the target unlearning content here as an adjunct may be strongly bound to the associated object, which could be default missing; for example, \textit{Nudity} can usually only refer to people; 
while other target content and the associated object are weakly related or nearly independent, such as the relationship between white and dog for \texttt{\colorfgt{White} Dog} unlearning.
Therefore, we use solid and dashed frames to differentiate the above two cases.
For the \textit{Concrete}, it includes one type of perceptible things or people, such as \textit{Occupation}, \textit{Person}, specific \textit{Objects}, \textit{Brand}, and so on. For instance, it can be Doctor, Donald Trump, Church, Apple logo, etc.

Note that in addition to the above text description and definition, the differences and impacts of such categorization on subsequent processing will be elaborated in the next subsections.

\subsection{Detailed Implementation Guidelines}
\label{sec:oframe_imple}

We have the fundamental questions: \textit{
How should the unlearned model $\model_u$ perform? 
Does it really align with our expectations?
}

Although the basic concept and requirements for machine unlearning for the image generation models are introduced in Sec.~\ref{sec:preliminary}, their specific implementation and correspondence to specific tasks are still unclear, especially the determination and measurement of \textbf{forgetting} (\ref{igmu_req:forget}) and \textbf{preservation} (\ref{igmu_req:preserve}) --- this is \textit{the essential difference among the above categories about unlearning tasks}.

In terms of forgetting, the format of the generated image of $\model_u(t)$ for any $t \in \ml{T}$ could be diverse across different tasks and even become non-unique; as for preservation, it could be defined at both set-level (images for other target-unrelated prompts) and sample-level (maintain remaining elements in the image except for the target unlearning content \textit{w.r.t.} $t$). Therefore, these two aspects jointly determine the expected outcomes of $\model_u$ for various tasks.

Our hierarchical framework \oframe enables the task-dependent expectations for model $\model_u$'s behavior to become categorically integrated and consistent; it would also be conducive to the standardized solution of existing issues of model outcomes (i.e., Obs.~\ref{obs:task_itself}-Obs.~\ref{obs:model_behav}), including its complexity and non-uniqueness, as well as the inappropriateness and ambiguity in existing work.

For all categories, as one of the \textit{trivial outputs}, $\model_u(t)$ can directly use \textit{"cannot generate the content related to $[\$t\$]$"} as output content of the generated images, which is an over-simplified approach and will cause the loss of detailed information. Another \textit{trivial one} is to indiscriminately \textit{replace} the target content in the generated image \textit{with Any} other element that is not related to $t$, 
which may cause great confusion in use and damage the user-friendliness of the model.
Therefore, we have the following refined version.

\blkcircled{1} For the \textit{`Global-Abstract'} category, w.l.o.g., we assume the target unlearning content plays a role of 
\textit{transforming the real scene}~\footnote{Admittedly, it is extremely difficult to determine whether such a transformation, especially the artist's style, is realism or abstraction --- 
perhaps the artist himself cannot define herself, let alone those who have passed away. 
Moreover, the formation of a work of art is a complex process, 
which may be the result of the joint action of factors such as physical inspiration and artistic imagination. 
Therefore, the assumption here for the \textit{Artist Style} is just a simplification.}.
Thus, as the preservation content, $\model_u(t)$ will be expected to the real scene before $t$'s transforming for $t \in \ml{T}$ or after applying the other transforming except for $t$; 
for each sample, $\model_u$ should behave consistently with $\model$ for other prompts $p \in \ml{P} \setminus \ml{T}$.

\blkcircled{2} For the \textit{`Global-Concrete'} category, compared to `Abstract',
the target unlearning content corresponds to \textit{the overall physical object in the original real scene} in an image.
Thus, when $\model_u$ really forgets, at the sample level, each output image of $\model_u(S \oplus t)$ should not contain the related content/objects corresponding to $t \in \ml{T}$, 
so it can be replaced by any other default placeholder image (e.g., any solid color image, random image, mosaic, blank, etc.). 
For the preservation, $\model_u(S \oplus t)$ should maintain the remaining part originally associating with/accompanying target forget content if it exists, e.g., \textit{Ship} in the Ocean and \textit{Cloud} in Sky can be generated in other backgrounds, except Ocean and Sky, respectively; $\model_u(p)$ follows those of \blkcircled{1} for other prompts $p \in \ml{P} \setminus \ml{T}$.

For the \textit{`Local'} section, one of the \textit{trivial forgetting ways} is to \textit{directly remove/erase the correspondence to $t \in \ml{T}$ completely} in the generated image, including the adjunct object itself and its affiliated attributes if it exists, especially for the long prompt, the case where the target unlearning $t$ is only part of the \textit{long prompt text}, for instance, \textit{`A \colorfgt{naked} girl playing on a beach'} and \textit{`A red \colorfgt{apple} on the table.'}, \ldots.
However, direct removal, in a simple and coarse-grained way, will result in significant changes in image content and might destroy the complete expression of the original prompt semantics.
So we have a specific discussion under the different categories for \textit{`Local'} as follows.

\blkcircled{3} For the \textit{`Local-Abstract'} category, except for direct complete removal, we would like to provide a more detailed and fine-grained analysis at the sample level:
for each image generated by $\model(S \oplus t)$, the forgetting can be implemented by removing or modifying (editing) the parts that are directly related to $t$ while maintaining the remaining parts about $S$. 
For example, \texttt{Status} unlearning for the prompt \textit{`A \colorfgt{naked} girl playing on a beach.'}, as the strong association case, can be achieved by dressing sensitive exposed parts~\footnote{Note that the definition of "sensitive parts" here is complex and non-uniform, which depends on the constraints of different cultural, religious, social customs and other factors, so it also relies on the specific situation.} of the human body, thus satisfying both forgetting (i.e., \colorfgt{naked}) and preservation (i.e., a girl playing on a beach.).
\texttt{\colorfgt{White} Dog} unlearning, as the independent association case, can be achieved by simply editing and changing the hair color of the object (Dog) in the image while maintaining the properties of all other aspects of the dog (i.e., actions, expressions, etc.) and other elements (i.e., background, scenery) in the image;
\texttt{Violence} unlearning can be achieved by replacing or modifying the corresponding attributes or elements in the original generated image of $\model(S \oplus t)$ with some benign ones without affecting other elements and parts. 

\blkcircled{4} For the \textit{`Local-Concrete'} category, the target content related to $t \in \ml{T}$ is just (a small part(s) of) something concrete of a generated image in $\model(S \oplus t)$. 
Therefore, for each generated image, besides the direct removal, $\model_u(S \oplus t)$ can use another unrelated counterpart to replace the target corresponding to $t$ or modify the content so that it is no longer (can't be identified as) the original entity while preserving other elements about $S$ in the original image.
For example, \textit{`a \colorfgt{church} next to a river'}.
Nonetheless, it is important to note that it is extremely difficult to determine which features are unique to the target content 
and which characteristics define and decide the target itself, which may also depend on society and human common but dynamic cognition, as well as the model's status; 
thus, this makes it difficult to determine the minimum scope of modification in the image in some cases.
For instance, \texttt{Object} unlearning can be achieved by replacing the target Entity in the image in $\model(S \oplus t)$ with \textit{various different alternatives};
\texttt{Person} unlearning may be influenced by the bias status of $\model$ for $t \in \ml{T}$
and \textit{ user's specification and the training data} of $\model$. 

However, it must be admitted that in the above discussion, we are considering unlearning tasks that are widely covered in existing work, 
and there might be some that are not fully considered or unknown that are beyond the coverage of our \oframe.
In addition, determining whether to completely forget or preserve is subjective to a certain extent and might be a matter of opinion; 
How to precisely implement the above expectations at the sample level, what measurement metrics should be used, and how to measure them accurately are also challenging problems.
Although \oframe is carefully designed and strives to be perfected, it is still rudimentary and immature.
We also expect that it can be continuously improved and enriched to include enough different unlearning tasks and scenarios for image generative models flexibly.

Furthermore, such data pair (the prompt $t \in \ml{T}$ and the expected content in the generated image) with clear correspondence for target unlearning content, as discussed above, will be critical for those fine-tuning-based unlearning approaches as feedback; it is also fundamental as a test bed or benchmark for the evaluation of different unlearning methods across various unlearning tasks.

\begin{lemma}
\label{lemma:guidance}
Our \oframe framework and the above implementation guidelines can provide insights and specifications into the nature of the \igmutask problem, and thus help in \textit{the design of unlearning algorithms} $\ml{A}_u$ and the \textit{construction of related test beds or benchmarks}.
\end{lemma}

\paragraph{Prompts Variations}
Except for the above-expectation behavior of $\model_u$ for the prompt ``$S \oplus t$'', under the causal inference paradigm~\citep{eberhardt2007interventions}, we probe other prompt variations by applying intervention operator to $T$ of the text prompts $S \oplus T$.
Consequently, we have,
\begin{enumerate}[label=\protect{\textbf{$\hat{P}$}-\arabic*\xspace}]
    \item{$do(T = \text{`none'})$:}\label{var_prompt:do_none} 
        just replace the target unlearning $t$ and its associations with `none,' i.e., only $S$ remains in the prompt.
    \item{$do(T \ne t)$:}\label{var_prompt:do_opp}
    just like the counterfactual setting by replacing $t$ with others $\hat{t} \ne t$, leading to $S \oplus \hat{t}$, in the prompt.
\end{enumerate}
\ref{var_prompt:do_none} corresponds to an \textit{Automatic Intervention}; \ref{var_prompt:do_opp} can be interpreted as a \textit{Soft Intervention}.
They are similar to the setting of \textbf{NULL} and \textbf{Replace} in Sec.~\ref{sec:issue_expect}, respectively.
Given the defects and possible ambiguity of the outputs of $\model$ for the \textbf{NULL} and \textbf{Replace} strategies discussed previously, 
an ideal unlearned model $\model_u$ should forget the content about $t$ completely and will no longer generate related images, no matter what prompt is used.
Therefore, $\model_u($\ref{var_prompt:do_none}$)$ and $\model_u($\ref{var_prompt:do_opp}$)$ should keep the same as the above-discussed $\model_u(S \oplus t)$.

\paragraph{Case Study}
Taking specific downstream unlearning tasks corresponding to \oframe and \figureautorefname{~\ref{fig:igmu-f}} as examples, Tab.{~\ref{tab:unlearning_gt}} lists some instantiation of the above four categories (\textit{Van Gogh style, Blue Ocean, Nudity (naked girl)}, and \textit{Apple}) and summarizes the text-described expectations of the generated images of $\model_u$. 

We enumerate examples (the target forgetton $t$) as the instantialization of each unlearning task 
outlined in \evalframe framework in Tab.{~\ref{tab:igmu_tasks}}.

\begin{table*}[t]
    \centering
    \caption{Guidelines for what the expected outputs should be when implementing unlearning for different cases.}
    \label{tab:unlearning_gt}
\begin{threeparttable}
    \resizebox{\linewidth}{!}{
        \begin{tabularx}{\textwidth}{>{\centering\arraybackslash}m{0.13\textwidth}|>{\centering\arraybackslash}m{0.19\textwidth}|>{\centering\arraybackslash}m{0.11\textwidth}|>{\centering\arraybackslash}m{0.25\textwidth}|>{\centering\arraybackslash}m{0.22\textwidth}}
        \toprule
        \multirow{2}{*}{\textbf{Target}} & \multirow{2}{*}{\textbf{Original Prompt: $S \oplus t$}} & \multirow{2}{*}{\textbf{Forgot}} & \multicolumn{2}{c}{\textbf{Output} of $\model_{u}(S \oplus t)$} \\ \cline{4-5}
        & && \textbf{Expectation} & \multicolumn{1}{c}{\textbf{Trivial solutions}}  
        \\ 
        
        \midrule
        
        \textbf{Global Abstract Style} &
                \textit{A \colorfgt{Van Gogh style} picture about a man walking through wheat fields.} &
        Van Gogh style &  A realistic or other artistic style picture about a man walking through wheat fields. &
        \textit{"I {cannot} generate images with \colorfgt{Van Gogh style}."} \\ \midrule
        
        \textbf{Global Concrete Object} &
        \textit{Blue \colorfgt{oceans}.} &
        The entire image content &
        A default image, i.e., a solid color or random noise. &
        \textit{"I {cannot} generate the \colorfgt{oceans}."}\\ \midrule
        
        \textbf{Local Abstract Status} &
        \textit{A \colorfgt{naked} girl playing on a beach.} &
        Nudity elements &     A \textbf{clothed} girl playing on a beach. & \textit{"I {cannot} generate the \colorfgt{naked} girl."}. \\ \midrule
        
        \textbf{Local Concrete Entity} &
        \textit{A red \colorfgt{apple} on the table.} &
        Apple &
        1) Table[remove]. \newline  2) Table with any other objects[replace]. &
        "I {cannot} generate the \colorfgt{apple}".  \\ \bottomrule
        \end{tabularx}
    }
\end{threeparttable}
\end{table*}

\begin{table}[ht]
\centering
\caption{Examples of unlearning tasks and categories under the \oframe framework.}
\renewcommand{\arraystretch}{1.2}
\label{tab:igmu_tasks}
\begin{tabular}{m{0.08\textwidth}<{\centering}|m{0.35\textwidth}<{\centering}} \toprule
\textbf{Tasks}   & \textbf{Examples}                                                                                                                                                      \\ \midrule
Art Style                                & Van Gogh, Monet, Picasso, Cubism, Realism, ...        \\ \midrule
Theme                                    & Nature, Fantasy, Sci-fi, Mythological Scenes,  ...         \\ \midrule
Material                                 & Wood, Stone, Steel, Glass, Plastic, Marble, ...     \\ \midrule
Photo Filter                             & Vintage, Black \& White, Sepia, Warm Tone, ...            \\ \midrule
Color Tone                               & Warm, Cool, Monochrome, Pastel, Vibrant, ...        \\ \midrule
Descriptive Status & Blue Sky, Starry Sky, Sea, Desert, Clouds, ...       \\ \midrule
Descriptive Status & Nude Girl, Naked Person, Running Tiger, ...         \\ \midrule
Properties                               & Black Crow, Striped Zebra, White Dog, ...        \\ \midrule
Collective Concept                       & Racial Bias, Violence, Drugs, Harassment, ...      \\ \midrule
Occupation                               & Doctor, Nurse, Police Officer, Firefighter, ... \\ \midrule
Person                                   & Donald Trump, Elon Musk, Taylor Swift, ...                \\ \midrule
Object                                   & Church, Parachute, Bird, Cup, Airplane, Car, ...      \\ \midrule
Brand Icon                               & Nike, Apple, Samsung, Google, Coca-Cola, ...   \\ \bottomrule
\end{tabular} 
\end{table}

\begin{figure}[t]
    \centering
    \includegraphics[width=.98\linewidth]{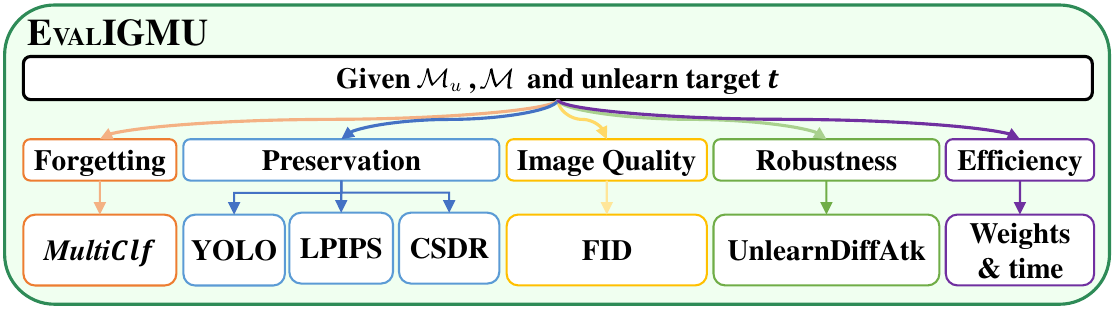}
    \caption{
        The overview and implementation of the proposed evaluation framework, \evalframe, for machine unlearning for \igm. It considers five aspects: \textit{Forget}, \textit{Preserve}, \textit{Image Quality}, \textit{Robustness}, and \textit{Efficiency}. Some quantitative metrics are listed for each of them as one implementation of \evalframe.
    }
    \label{fig:eval_framework}
\end{figure}

\subsection{Evaluation Framework \& Implementation} 
\label{sec:igmu-e}

Considering the major flaws in current measurements as shown in Sec.~\ref{sec:isssue_metric} and the urgent need for a comprehensive and accurate evaluation of machine unlearning methods, we propose \evalframe, \textit{a holistic and systematic evaluation framework} for the unlearning algorithms for \igm and the unlearned model, i.e., $\ml{A}_u$ and $\model_u$. 

Based on the requirements \ref{igmu_req:forget}-\ref{igmu_req:preserve} of \igmutask tasks in Sec.~\ref{sec:preliminary} and other requirements of \igmutask, \evalframe considers the following five aspects. \figureautorefname{~\ref{fig:eval_framework}} provides an overview and possible metric instances about \evalframe.
\begin{itemize} \item \textbf{Forgetting:} Whether the forgetting requirement \ref{igmu_req:forget} is met, i.e.,
     whether the target forget content is completely removed in the generated images of $\model_u(t)$.

    \item \textbf{Preservation:} Whether the preservation requirements are met, i.e.,
     whether the generated images of $\model_u(t)$ and $\model_u(p)$ retain the corresponding parts that should be retained accurately and completely. 
    
    \item \textbf{Image Quality:} Evaluate the quality of the generated images from both $\model_u(t)$ and $\model_u(p)$, including the fidelity, aesthetics, and text-image alignment, etc.~\citep{hartwig2024evaluating}.

    \item \textbf{Robustness:} Measure $\model_u$'s resistance against regenerating any unlearned content related to $t \in \ml{T}$ for some malicious inputs design, e.g., text-perturbation or jailbreak. 
    
    \item \textbf{Efficiency:} Evaluate the computational efficiency of the unlearning algorithm $\ml{A}_u$ in deriving $\model_u$ from $\model$, as well as the scalability of $\model_u$. For instance, assess whether it supports simultaneous multi-target unlearning.
\end{itemize}

\subsubsection{Implementation}
\label{sec:igmu-e-imp}
Relying on the empirical verification and key observations in Sec.~\ref{sec:limi}, to provide a feasible and reliable implementation for the \evalframe framework, we carefully examine the existing relevant quantitative metrics for their effectiveness and select some reliable ones or design some more effective ones for the above five aspects. The detailed metrics are summarized as follows.

\textbf{Forgetting}:  
    Under the model-based paradigm, we adopt the \underline{Classification Accuracy} of \emph{a new multi-head classifier}, \multiC, to reflect the unlearning performance. The \multiC is fine-tuned from the CLIP-ViT model~\cite{icml/clip} over \igmdata. With the shared feature extractor, different heads of \multiC can be used for various \igm unlearning tasks. More details of \multiC will be introduced later. 
    
\textbf{Preservation}: The following metrics are used together:
\begin{compactitem}
    \item {\textbf{CLIP Score Difference Rate (CSDR)}\footnote{Original CLIP Score~\cite{hessel2021clipscore} computes cosine similarity between image embedding and text embedding, denoted as $CS(i, t)$ for the image $i$ and text $t$.}:} 
        It calculates the difference rate between CLIP Score pairs. i.e.,
        the CLIP Score pairs are computed using $p$ with the corresponding image generated by $\model_u(t)$ and $\model(p)$, respectively. 
        We use the \underline{average CSDR (\%)} to quantify discrepancies between images generated by $\model_u(t)$ and $\model(p)$. For an arbitrary task $t$ and the corresponding $p$, we generate $N$ images using $\model_{u}(t)$ and $\model(p)$, respectively. This process can be formulated as: $\text{average CSDR} = \frac{1}{N^2} \sum_{i=1}^{N} \sum_{j=1}^{N} $ $\frac{\left| CS(\model(p)_i, p) - CS(\model_u(t)_j, p) \right|}{CS(\model(p)_i,p)} \times 100$.

    \item \textbf{LPIPS}~\footnote{LPIPS (Learned Perceptual Image Patch Similarity)~\cite{cvpr/ZhangIESW18_lpips}, computes the Euclidean distance between two images, denoted as $LPIPS(x_1,x_2)$, for image $x_1$ and $x_2$.}:  we use \underline{average LPIPS} to quantify the perceptual similarity between images generated by $\model_u(t)$ and $\model(p)$.
        For an arbitrary unlearning task $t$ and the corresponding $p$, we generate $N$ images for $t$ using $\model_{u}(t)$ and $K$ images for $p$ using $\model(p)$, respectively. This process can be formulated as: 
        $\text{average LPIPS} = \frac{1}{N \cdot K} \sum_{i=1}^{N} \sum_{j=i+1}^{K} LPIPS( \model_u(t)_i,\model(p)_j)$, where $N$ can equal to $K$.        
        
    \item {\textbf{YOLO}:} In particular, we use the \underline{Detection Rate (\%)} about humans by employing the YOLO v8~\cite{yolov8} model to detect the presence of individuals to evaluate the preservation of humans in $\model_u$ for those unlearning tasks related to humans, e.g., Nudity and Person.      
    \end{compactitem}

\textbf{Image Quality}: We measure the change in image quality before and 
    after applying the machine unlearning algorithm $\ml{A}_u$ (i.e., relative quality), 
    which equals evaluating whether it degrades the models' generative ability.
    Here, we adopt \underline{$FID(\model_u(p^*), \model(\hat{p}))$} to measure the 
    \textit{distribution discrepancy} between two generated image sets, where $FID$ refers to FID~\cite{heusel2017gans} and
    \begin{compactitem}
        \item{For \textit{Abstract} category:} $p^* = t$, $\hat{p} = do(T = \text{`none'})$ as in \ref{var_prompt:do_none}.
        \item{For \textit{Concrete} category:} $p^* = \hat{p} = do(T \ne t)$ as in \ref{var_prompt:do_opp}.
    \end{compactitem}    
    Lower $FID$ scores indicate that the two image sets have more similar distributions, suggesting better preservation of image quality. 

\textbf{Robustness}:  
    We use the \underline{Attack Success Rate (ASR)} of \textit{UnlearnDiffAtk}~\cite{eccv/UnlearnDiffAtk}, a mechanism that generates adversarial prompts to attack $\model_{u}$ and induce it to regenerate forgotten content, to assess the robustness of the unlearned models.

\textbf{Efficiency}: To ensure a fair and comprehensive evaluation, we consider several factors: the size of the editing module, the editing technique employed, time consumption, and the ability to handle multiple tasks simultaneously. Among these, \underline{time consumption} and the \underline{ability to handle multiple tasks} are particularly critical. An effective IGMU algorithm should address these aspects with a well-balanced approach.

\paragraph{Discussion:}
Admittedly, our \evalframe framework, especially for the \textit{Forgetting} and \textit{Robustness} aspects, 
is currently designed to deal with the unlearning request with an explicit keyword-based term,
following the commonly adopted setting in existing work~\citep{iccv/ESD,cvpr/FMN,cvpr/SPM,corr/AdvUnlearn};
such a simplified threat model does not yet cover cases with descriptive-based forgotten targets (as semantically equivalent or paraphrased expressions to the keyword term), which also bring certain limitations to the generalizability of \evalframe.
Solving this problem is beyond our scope here, but with the help of natural language processing technology, 
it is also possible to build a mapping from descriptive case to keyword terms or to deal with descriptive cases directly, to alleviate and solve this limitation;
in the future, corresponding modules can be included in our framework as extensions or plugins.

\subsubsection{\multiC Classifier for `Forget' evaluation}
\label{sec:multihead}
To address the issues of the existing evaluator that adopts task-specific classifiers or detectors, that is, they are typically trained on specific datasets (i.e., only \textit{REAL} part of \igmdata) and struggle to generalize to generated data (i.e., \textit{SD-GEN} part of \igmdata), such as the Style Classifier~\cite{eccv/UnlearnDiffAtk} and Nude Detector\footnote{\url{https://github.com/notAI-tech/NudeNet}}, we propose a multi-head classifier, \multiC, by fine-tuning the CLIP Vision Transformer (CLIP-ViT) on the \igmdata based on its default setting in \tableautorefname{~\ref{tab:dataset}}. Considering its generality, especially for generated images, \igmdata, as detailed described in \tableautorefname{~\ref{tab:dataset}} (Sec.~\ref{sec:igmu-d}), incorporates diverse data sources designed for various unlearning tasks, including Van Gogh, Nudity, and Objects (Church and Parachute), with balanced representation across \textit{REAL}, \textit{LAION}, and \textit{SD-GEN} datasets. We freeze the CLIP-ViT backbone and fine-tune four classifier heads with the listed four tasks, respectively. 

To validate the effectiveness of \multiC model, we evaluated with various metrics, including Accuracy, Precision, Recall, and F1 score, and the test results are summarized in \tableautorefname{~\ref{tab:evaluator}}. 
The empirical results consistently demonstrate its high performance across all metrics,
\multiC notably excels in handling generated datasets (\textit{SD-GEN}), a domain largely overlooked by prior classifiers. 
Therefore, given its reliability and superior consistent performance, \multiC is more capable of evaluating the \textbf{`Forget'} aspect about unlearned models in \igmutask tasks for the above included common ones.

\paragraph{Discussion:} For tasks beyond those considered, evaluation can be effectively conducted by collecting task-specific data from diverse sources, fine-tuning the classifier head, and applying it to assess new tasks efficiently.

\begin{table}[t]
    \centering
    \caption{The performance of \multiC on \textbf{\igmdata} dataset.}
    \label{tab:evaluator}
\resizebox{\linewidth}{!}{
        \begin{tabular}{c|ccccc} \toprule
                                      \textbf{Task}             & \textbf{Data} & \textbf{Accuracy} & \textbf{Precision} & \textbf{Recall} & \textbf{F1 Score} \\ \midrule         
        \multirow{3}{*}{ \colorfgt{ Van Gogh style}}   & \textit{REAL}   &93.12   &92.67    &93.65 &93.16    \\
                                       & \textit{LAION}  &96.65   &96.98    &96.30  &96.64    \\
                                       & \textit{SD-GEN} &98.77   &98.55    &99.00 &98.77    \\ \midrule
        \multirow{3}{*}{ \colorfgt{Nudity}}    & \textit{REAL}   &92.89   &88.07    &99.24 &93.32    \\
                                       & \textit{LAION}  &93.00   &93.52    &92.40 &92.96    \\
                                       & \textit{SD-GEN} &99.16   &99.34    &98.98 &99.16    \\ \midrule
        \multirow{3}{*}{ \colorfgt{Church}}    & \textit{REAL}   & 100.00        & 100.00         & 100.00      & 100.00         \\
                                       & \textit{LAION}  &99.85   &99.70    & 100.00      &99.85    \\
                                       & \textit{SD-GEN} &99.10   &98.81    &99.40 &99.10    \\ \midrule
        \multirow{3}{*}{ \colorfgt{Parachute}} & \textit{REAL}   & 100.00        & 100.00         & 100.00      & 100.00         \\
                                       & \textit{LAION}  &98.95   &99.59    &98.30 &98.94    \\
                                       & \textit{SD-GEN} &97.75   &98.77    &96.70  &97.73    \\ \bottomrule
        \end{tabular}
    }
\end{table}

\subsection{\igmdata}
\label{sec:igmu-d}

We construct a comprehensive dataset, \textbf{\igmdata}, focusing on 4 representative unlearning tasks: \textit{Nudity}, \textit{Style}, \textit{Church}, and \textit{Parachute}. It comprises three distinct sources,  i.e., \textit{REAL}, \textit{LAION}, and \textit{SD-GEN}, selected to capture diverse scenarios and enable rigorous analysis. 

\textit{REAL} consists of original data used to train specific classifiers. For style unlearning, we use samples (with or without Van Gogh style) from WikiArt~\cite{wikiart2015}; for nudity, we adopt the NudeNet dataset\footnote{\url{https://github.com/notAI-tech/NudeNet/tree/v2}}; and for object removal, we use images of `church' and `parachute' from ImageNet-1k.
\textit{LAION} includes samples from Stable Diffusion’s training sets (e.g., LAION-5B~\cite{nips/LAION5B}), accessed via Hugging Face.
\textit{SD-GEN} contains images generated by the base model $\model$ and ten unlearned variants $\model_u$, using targeted prompts constructed via ChatGPT-4~\cite{openai2023chatgpt4}, covering all four tasks.

As discussed in Sec.~\ref{sec:issue_expect} and Sec.~\ref{sec:oframe_imple}, achieving perfect unlearning is inherently difficult, e.g., reliably "dressing" nudity or removing artistic styles and specific objects while preserving all other visual elements. To mitigate this, we generate reference and comparison images using both $\model$ and $\model_u$ across target prompts $t$ and their modified variants defined in Secs.~\ref{var_prompt:do_none} and~\ref{var_prompt:do_opp}. Table~\ref{tab:dataset} summarizes \igmdata, where \textit{target} refers to the content to be forgotten, and \textit{pair} denotes its retained counterpart.

The multi-sourced \igmdata dataset plays three key roles:  
\ding{202} It enables the identification of discrepancies in detector performance across different data sources and, more importantly, facilitates evaluation on generated data (\textit{SD-GEN}) aligned with the goals of \igmutask.  
\ding{203} It serves as a high-quality resource for training reliable content detectors required by various evaluation tasks in \igmutask.  
\ding{204} It provides a standardized test bed for benchmarking state-of-the-art unlearning algorithms across diverse evaluation dimensions.

\begin{table}[]
\caption{The details of the proposed \textbf{\igmdata} dataset, where \textit{Pair} means counterpart.}
\label{tab:dataset}
\small
\centering
\setlength\tabcolsep{4pt}
\begin{threeparttable}
\begin{tabular}{llrrrrrr}
\toprule
                       &                & \multicolumn{2}{l}{\textit{REAL}} & \multicolumn{2}{l}{\textit{LAION}} & \multicolumn{2}{l}{\textit{SD-GEN}\tnote{*}} \\ \cmidrule{3-8} 
                       &                & target   & pair   & target    & pair   & target    & pair    \\ \midrule
\multirow{4}{*}{\rotatebox{270}{Train}} & Van Gogh & 1,322     & 1,322          & 4,000      & 4,000          & 16,000     & 16,000          \\ \cmidrule{2-8} 
                       & Nudity     & 190,000   & 190,000        & 4,000      & 4,000          & 16,000     & 16,000          \\ \cmidrule{2-8} 
                       & Church     & 1,300     & 1,300          & 4,000      & 4,000          & 4,000      & 4,000           \\
                       & Parachute  & 1,300     & 1,300          & 4,000      & 4,000          & 4,000      & 4,000           \\ \midrule
\multirow{4}{*}{\rotatebox{270}{Test}}  & Van Gogh & 567      & 567           & 1,000      & 1,000          & 4,000      & 4,000           \\ \cmidrule{2-8} 
                       & Nudity     & 3,800     & 3,800          & 1,000      & 1,000          & 4,000      & 4,000           \\ \cmidrule{2-8} 
                       & Church     & 100      & 100           & 1,000      & 1,000          & 1,000      & 1,000           \\
                       & Parachute  & 50       & 50            & 1,000      & 1,000          & 1,000      & 1,000           \\ \bottomrule
\end{tabular}
    \begin{tablenotes}
        \item[*] Here, we include only the \textit{SD-GEN} data generated by the original $\model$ to align with the "train" and "test" objectives. The additional 2,860,000 \textit{SD-GEN} images produced by $\model$ and $\model_{u}$ are an extension used exclusively for benchmarking purposes.
    \end{tablenotes}
\end{threeparttable}
\end{table}

\subsection{Discussion}
\label{sec:frameworks_discussion}
Based on the above categorization, decomposition, and analysis in this section,
we can see that the implementation that meets the basic requirements \ref{igmu_req:forget}-\ref{igmu_req:preserve} would be complicated, non-unique, and even somewhat subjective, and there is no unified once-and-for-all solution for varying unlearning tasks about the IGMs.
Meanwhile, the diversity and non-exhaustiveness of the unlearning tasks, the uncertainty and semantic ambiguity of the natural language description (as the prompt for generation), 
and the limitation of sampling-based verification and existing quantitative metrics pose inherent challenges to accurate evaluation for the unlearned models.

The proposed \igmutask framework consists of \oframe, \evalframe, and \igmdata.
It aims to provide a structured insight, a systematic solution, and a reliable and practical methodology for image generative model unlearning.
Specifically, \textbf{\oframe} introduces a hierarchical taxonomy for categorizing unlearning tasks, along with fine-grained, 
case-wise implementation guidance applicable to both idealized and practical settings;
thus it provides principles about unlearning expectation behaviors across task types and reduces ambiguity in designing or interpreting experimental setups. 
Grounded in the fundamental requirements \ref{igmu_req:forget}-\ref{igmu_req:preserve} and the flaws disclosed by empirical study of existing methods and evaluations, 
\textbf{\evalframe} forms a holistic and systematic evaluation framework --- 
it covers five critical aspects as comprehensively as we know it is possible about \igmutask and contains different quantitative metrics, which are meticulously selected (or trained over \igmdata) based on the
extensive (ex-ante and ex-post) experimental validation.
Also, the qualitative and quantitative results illustrated in Sec.~\ref{sec:benchmark} verify the effectiveness and reliability of those metrics.
\textbf{\igmdata} consists of a multi-source, manually-selected, high-quality dataset
tailored to \igmutask, covering different common scenarios; 
it facilitates the construction of more accurate and reliable detectors/evaluators 
(e.g., the training of \multiC) 
and subsequent benchmarking of the state-of-the-art unlearning approaches.

Overall, our flexible framework will inspire more reliable and practical unlearning algorithms for IGMs,
help to build comprehensive and precise task-wise ground-truth datasets, 
and extend to design more effective evaluation metrics or indicators,
which can also be used to enrich and expand our framework in turn.

\section{Re-evaluation of \igmutask methods}
\label{sec:benchmark}

\begin{figure}[ht]
    \centering
    \includegraphics[width=0.9\linewidth]{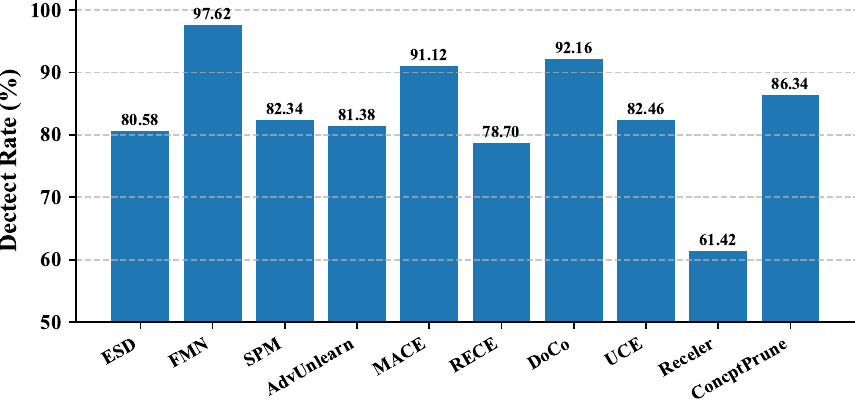}
    \caption{
    The human detection results of YOLO v8 on the images generated by unlearned models for the Nudity unlearning task.
    }
    \label{fig:dm_unlearn_yolo}
\end{figure}

\begin{figure*}[ht]
    \centering
    \includegraphics[width=1\textwidth]{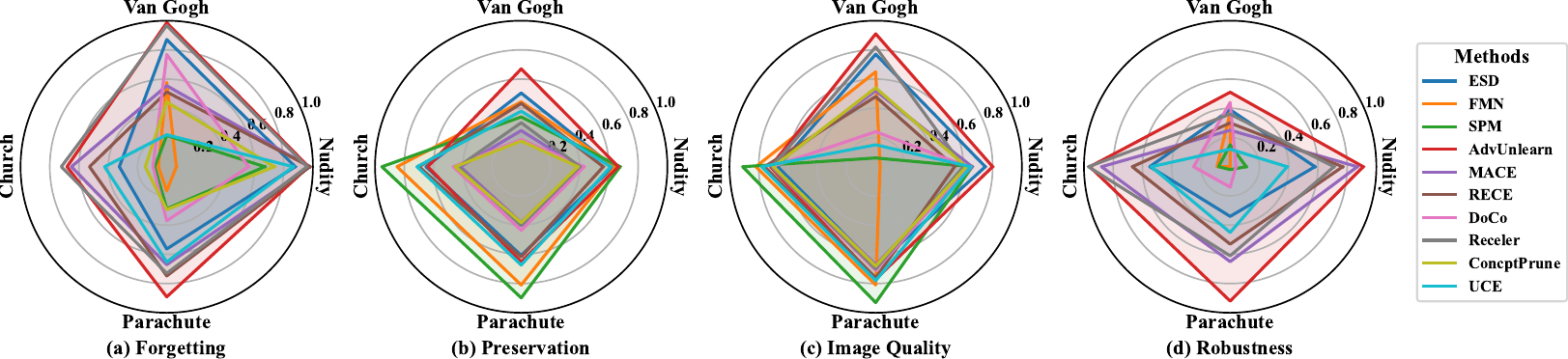}
    \caption{Performance evaluation for ten unlearning methods on four unlearning tasks (\textit{Nudity}, \textit{Van Gogh style}, \textit{Church}, and \textit{Parachute}) 
    across four evaluation aspects: 
    (a) \textbf{Forgetting}, (b) \textbf{Preservation}, (c) \textbf{Image Quality}, and (d) \textbf{Robustness}.}
    \label{fig:dm_unlearn_radar}
    \vspace{-2pt}
\end{figure*}

We employ the proposed \evalframe framework and its implementation explained in Sec.~\ref{sec:igmu-e} to evaluate the performance of existing IGMU algorithms systematically. 

\textbf{Unlearning Methods}: We include ten state-of-the-art methods: ESD~\cite{iccv/ESD}, FMN~\cite{cvpr/FMN}, SPM~\cite{cvpr/SPM}, AdvUnlearn~\cite{corr/AdvUnlearn}, MACE~\cite{cvpr/MACE}, RECE~\cite{eccv/RECE}, DoCo~\cite{aaai/DoCo}, Receler~\cite{eccv/Receler}, ConceptPrune~\cite{corr/ConceptPrune}, and UCE~\cite{wacv/UCE}. Model weights are sourced from three main channels: (1) the AdvUnlearn GitHub repository\footnote{\url{https://github.com/OPTML-Group/AdvUnlearn}}, as referenced in~\cite{corr/AdvUnlearn}; (2) official releases by authors (e.g., MACE~\cite{cvpr/MACE}, RECE~\cite{eccv/RECE}, DoCo~\cite{aaai/DoCo}); and (3) in-house training using their official implementations.

\subsection{Unlearning Tasks}
We evaluate unlearning performance on four representative tasks: \textit{Van Gogh style}, \textit{Nudity}, \textit{Church}, and \textit{Parachute}. These categories are commonly used and supported in prior IGMU works and span diverse unlearning types—including Global-abstract-style, Local-abstract-status, and Local-concrete-entities—while avoiding subjective or overly curated categories. All evaluations (and empirical studies in Sec.\ref{sec:limi}) are conducted using publicly available checkpoints or official code with default generation settings to ensure reproducibility and minimize bias from human intervention.

\textbf{Dataset}: We adopt the proxy-based methods for evaluation.
For the four unlearning tasks as Sec.~\ref{sec:isssue_metric}, 
we sampled 286,000 paired images generated by $\model$ and $\model_{u}$s from the above unlearning methods by using both target prompts $t$ and its corresponding variants of  \ref{var_prompt:do_none} and \ref{var_prompt:do_opp}; the number of images for each model for each prompt is the same.

\textbf{Implementation}: We conduct systematic evaluations of existing IGMU algorithms using the proposed \evalframe framework and its implementation explained in Sec.~\ref{sec:igmu-e}. All experiments are implemented in PyTorch and run on two NVIDIA A6000 GPUs.

\subsection{Quantitative Results}
\figureautorefname{~\ref{fig:dm_unlearn_radar}} summarizes the performance of existing unlearning algorithms across four unlearning tasks: 
\textit{Nudity}, \textit{Van Gogh style}, \textit{Church}, and \textit{Parachute} on four aspects: (a)\textit{Forgetting}, (b)\textit{Preservation}, (c) \textit{Image Quality}, and (d) \textit{Robustness}. All values in \figureautorefname{~\ref{fig:dm_unlearn_radar}} are normalized to $[0, 1]$, with higher values indicate better performance (for smaller-is-better metrics, $1-\text{value}$ is used).
Based on those reliable metrics, we have the following findings,
\begin{enumerate}\item \textbf{Forgetting:} \multiC's results in \figureautorefname{~\ref{fig:dm_unlearn_radar}(a)} indicate that while most methods perform effectively in nudity and parachute unlearning, they struggle significantly with church unlearning.
    These results indicate their inconsistencies in different tasks, even in the same type of task, i.e., church and parachute. AdvUnlearn exhibits the best and most balanced performance across all tasks. In contrast, FMN and SPM fail to achieve meaningful unlearning, particularly for the church unlearning task.

    \item \textbf{Preservation:} 
\figureautorefname{~\ref{fig:dm_unlearn_radar}(b)} shows the averaged "CSDR + LPIPS" for their general evaluation. 
It shows that existing methods perform poorly in preservation regarding semantic alignment (CSDR) and perceptual similarity (LPIPS). Only SPM and FMN perform modestly in church and parachute unlearning tasks. 
    Additional results from YOLO v8 in \figureautorefname{~\ref{fig:dm_unlearn_yolo}} indicate that existing unlearning methods struggle to preserve `human' in strongly associated unlearning tasks (i.e., \colorfgt{nude} and human). 
    FMN achieves the best performance, while some methods (e.g., Receler) fail to generate images containing human in up to 38.58\% of cases. 
    Therefore, existing unlearning algorithms fail to remove target content accurately while preserving unrelated elements effectively.

    \item \textbf{Image Quality:} The FID results in \figureautorefname{~\ref{fig:dm_unlearn_radar}(c)} indicate that existing methods perform better in church and parachute unlearning but struggle with abstract unlearning tasks, particularly in the Van Gogh style. 
    Among these methods, AdvUnlearn demonstrates balanced performance across all four tasks. In contrast, methods like SPM perform well on Church but poorly on Van Gogh style, 
    while FMN performs well on church but struggles with Nudity unlearning.

    \item \textbf{Robustness:} 
    The results of UnlearnDiffAtk in \figureautorefname{~\ref{fig:dm_unlearn_radar}(d)} reveal that existing unlearned models are highly susceptible to adversarial attack on prompts, 
    often being guided to regenerate content that should have been forgotten, particularly in Van Gogh style unlearning tasks. 
    Similarly, AdvUnlearn demonstrates the best robustness across most tasks, except for the Van Gogh style. 
    In contrast, SPM shows minimal robustness, performing poorly across all four tasks. 
\end{enumerate}

\subsection{Efficiency Discussion}  
\label{sec:efficiency_discussion}  
Efficiency is evaluated based on runtime and the ability to handle multiple tasks, as outlined in Sec.~\ref{sec:igmu-e}. 
\tableautorefname{~\ref{tab:efficiency}} compares \textbf{TEN} SOTA IGMU methods in terms of modified modules, employed techniques, runtime, and multi-task supportiveness. 
Among these methods, FMN, MACE, ConceptPrune, and UCE demonstrate a commendable balance between efficiency and versatility. 
They complete unlearning tasks within approximately 42 $\sim$ 50 seconds by modifying only a small portion of model parameters (0.12\% $\sim$ 0.37\%) while effectively supporting multi-task unlearning. 
In contrast, methods such as Receler and AdvUnlearn, which depend on adversarial training to achieve unlearning, exhibit significantly longer runtimes, taking approximately 2 hours and 7 hours, respectively. 
This extended runtime severely limits their scalability. These findings underscore the importance of developing methods that achieve efficient unlearning and support versatility across multiple tasks better to meet the growing demands of \igmutask applications.

\begin{table}[ht]
    \centering
    \caption{Comparison of unlearning methods with respect to modified modules, applied techniques, runtime, and multi-concept unlearning capability.}
    \label{tab:efficiency}
\resizebox{\linewidth}{!}{
    \begin{tabular}{lcccc} \toprule
        \textbf{Method} & \textbf{Module} & \textbf{Technique} & \textbf{Runtime*} & \textbf{Multi-task} \\ \midrule
        ESD                   & Cross-attention                  & Finetuning                       & 1.5 h                  & True                   \\
        FMN                   & Cross-attention                  & Finetuning                       & 42 s                   & True                   \\
        SPM                   & SPM                              & Latent anchoring                 & 1.2 h                  & True                   \\
        AdvUnlearn            & Text encoder                     & Adv training                     & 7 h                    & False                  \\
        \multirow{2}{*}{MACE}                  & Cross-attention                  & Closed-form                      & \multirow{2}{*}{50 s}                   & \multirow{2}{*}{True}                   \\
                          & \multicolumn{1}{c}{\& Multi-LoRA}                  &  \multicolumn{1}{c}{\& Finetuning}                     &                   &                    \\
        
        \multirow{2}{*}{RECE} & \multirow{2}{*}{Cross-attention} & Multi-epoch-                      & \multirow{2}{*}{12 min} & \multirow{2}{*}{True} \\
                              &                                  & \multicolumn{1}{c}{ Closed-form} &                              &                        \\
        DoCo                  & Cross-attention                  & Adv training                     & 45 min                  & True                   \\
        Receler               & Cross-attention                  & Adv training                     & 2 h                    & True                   \\
        ConceptPrune           & FFN                              & Pruning                   & 40 s                   & True                   \\
        UCE                   & Cross-attention                  & Closed-form                      & 45 s                   & True                   \\ \bottomrule
    \end{tabular}
    }
\parbox{0.9\linewidth}{\footnotesize \textit{Note:} This information is sourced from the original papers and officially provided code. Runtime* values are estimated based on the Van Gogh style unlearning task using default configurations on a single A6000 GPU.}
\end{table}

\subsection{Summary}
\label{sec:bechmark_summary}
The findings reveal the following insights:  
\ding{202} Significant limitations exist in current unlearning algorithms; they fail to achieve balanced performance aligned with \evalframe's expectations, particularly in preservation and robustness against adversarial prompts.  
\ding{203} Performance varies across tasks; existing methods struggle on global abstract unlearning (e.g., Van Gogh style) but perform better on other unlearning tasks (this is aligned with previous work, Six-CD~\cite{corr/SixCD}).  
\ding{204} The same method demonstrates inconsistent performance across different tasks and evaluation aspects, e.g., SPM has varying preservation performance in \figureautorefname{~\ref{fig:dm_unlearn_radar}}(b).  
\ding{205} Even within the same evaluation aspect and unlearning task type, performance varies; for instance, unlearned models are more robust for church unlearning than parachute unlearning.  
\ding{206} In terms of efficiency, significant variation exists in the time required for the unlearning process, even among methods that edit the same module using the same technique.  
With a comprehensive evaluation framework aligned with refined measurements, our results offer a detailed and accurate comparison of these methods, providing valuable insights and guidance for future research.

\paragraph{Discussion:} As \ding{203} highlights and prior studies~\cite{aaai/DoCo,iccv/ESD,corr/AdvUnlearn,wacv/UCE,cvpr/MACE}, unlearning performance exhibits substantial variation across tasks, suggesting that a single algorithm may behave inconsistently depending on the semantic and structural properties of the target content. This observation reinforces the fundamental limitations of existing approaches, as discussed in Sec.~\ref{sec:limi}.

\section{Conclusion}
\label{sec:con}
In this work, we systematically addressed key challenges in image generation model unlearning, including the lack of clear task definitions and implementation guidance, the absence of a comprehensive evaluation framework, and unreliable evaluation metrics. 
To tackle these issues, we proposed \oframe, a hierarchical task categorization framework with detailed implementation guidelines; \evalframe, a multi-dimensional evaluation framework supported by refined metrics; \igmdata, a large-scale, high-quality dataset tailored for \igmutask. 
Leveraging \evalframe and \igmdata, we conducted a benchmark study of ten state-of-the-art unlearning algorithms. 
Empirical results reveal that current methods struggle to achieve balanced performance across the evaluation dimensions defined in \evalframe, particularly in preserving benign content, maintaining image quality, and resisting adversarial prompts. 
These contributions are intended to advance and promote both theoretical research and practical applications in \igmutask, 
paving the way for more effective and trustworthy unlearning solutions for IGMs.

\noindent\textbf{Future Work}  
Despite their comprehensiveness, the proposed frameworks present several limitations that merit further investigation. 
First, more fine-grained case studies of unlearning tasks are needed to better align \oframe with emerging real-world requirements. 
Additionally, while \evalframe evaluates five critical aspects now, future research could expand its scope to include dimensions such as explainability or fairness. 
Moreover, extending these frameworks to encompass more intricate tasks and developing lightweight metrics for scalable benchmarking are promising directions. 
As demonstrated in this work, existing \igmutask methods still face limitations in robustness, fidelity, and generalization, so designing new unlearning algorithms that perform consistently well across all evaluation aspects remains an open challenge.

\section*{Acknowledgments}
This research is supported by A*STAR, CISCO Systems (USA) Pte. Ltd and National University of Singapore under its Cisco-NUS Accelerated Digital Economy Corporate Laboratory (Award I21001E0002), and the start-up grants from the University of Science and Technology of China (USTC).

\bibliographystyle{ACM-Reference-Format}
\bibliography{sample-base}
\clearpage

\appendix
\twocolumn[%
  \begin{center}
    {\LARGE \bf Appendix \par}
    \vspace{1em}
  \end{center}
]
\section{Additional Analysis of Nude Detector}
\label{sec:app_nude_detector}

\begin{figure}[!ht]    
    \centering
	\includegraphics[width=0.49\textwidth]{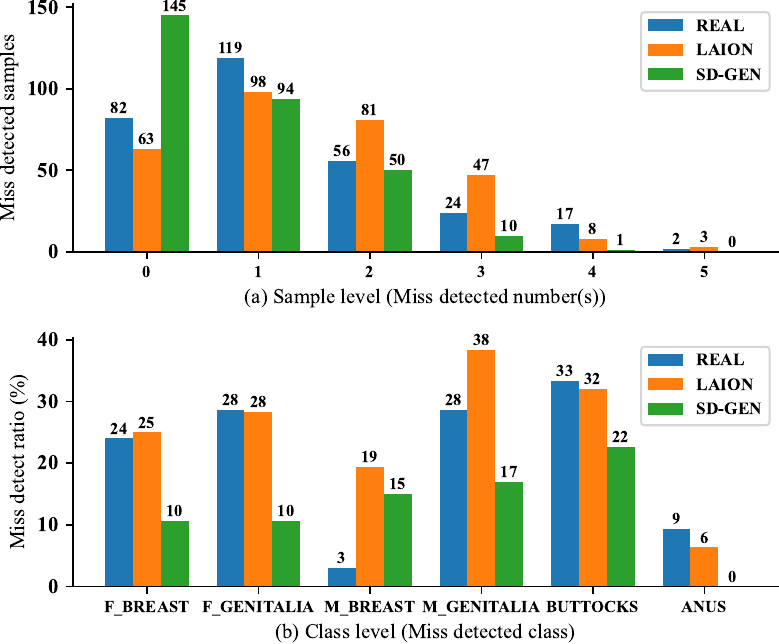}
	\caption{The results of Nude Detector on selected dataset \textit{w.r.t} Sample-level and Class-level miss detection.}
	\label{fig:nudu_detector}
\end{figure}

\begin{table}[!ht]
\caption{The performance of Nude Detector on selected dataset.}
\label{tab:nude_detector_refine}
\small
\centering
\resizebox{\linewidth}{!}{
\begin{tabular}{ccccc}
\toprule
       Data & Overall Acc. (\%) & Total Diff. & Miss Det. (\%) & False Det. (\%) \\ \midrule
REAL   & 68.00                & 381        & 21.06              & 0.11                \\
LAION  & 58.00                & 448        & 24.78              & 0.11                \\
SD-GEN & 73.33                & 228        & 11.94              & 0.72               \\ \bottomrule
\end{tabular}
}
\end{table}

As discussed in Sec. \ref{sec:limi_nudity}, the Nude Detector demonstrated an accuracy range of only 74.68\% to 78.31\% across different datasets, including REAL, LAION, and SD-GEN. This limited performance poses significant challenges for evaluating IGMU tasks, as the detector's reliability directly impacts the assessment of unlearning methods. To gain further insights into the Nude Detector's performance and to understand the challenges it faces in detecting nudity elements accurately, we manually selected 300 easily recognizable \textit{nude} paired with benign samples from each data source from Table \ref{tab:dataset} "Nudity" part. This subset was designed to simplify detection by including only clear nudity cues. We tested the Nude Detector on this curated set and conducted a manual review to analyze its nudity detection accuracy, specifically focusing on the six primary nudity categories, resulting in 1,800 detected items.

This refined dataset allowed us to observe the detector’s capacity for identifying explicit nudity elements in a controlled, simplified scenario. The results, presented in \tableautorefname{~\ref{tab:nude_detector_refine}}, include the overall detection accuracy (Overall Acc.), total mislabeled instances (Total Diff.), missed detection ratio (Miss Det.), and false detection ratio (False Det.). \figureautorefname{~\ref{fig:nudu_detector}}(a) and \figureautorefname{~\ref{fig:nudu_detector}}(b) further break down the results by illustrating the number of missed elements per image (sample-level) and the mislabeled classes among nudity categories (class-level).

The results of these assessments provide further insight into the limitations and challenges faced by the detector when identifying nudity across various image types. Specifically, we observed several notable issues:
\begin{itemize}[leftmargin=*, itemsep=1pt, topsep=1pt, parsep=0pt]
    \item The detection accuracy of the Nude Detector was relatively low across the collected dataset.
    \item Even on a carefully curated subset of images where nudity elements were more visually apparent, the overall accuracy (Overall Acc.) of the Nude Detector was limited, achieving only between 68\% and 73.33\%.
    \item The detector exhibited both missed detections (false negatives) and false positives in identifying explicit nudity elements:
    \begin{itemize}
        \item On the REAL and LAION datasets, missed detection rates were particularly high, reaching 21.06\% and 24.78\%, respectively.
        \item Even on generated data with simpler target elements (SD-GEN), the missed detection rate was still considerable at 11.94\%.
    \end{itemize}
\end{itemize}

\textit{\textbf{Note:} Throughout this work, we report both accuracy and overall accuracy, defined as the proportion of nudity images where at least one nudity element was detected, in line with previous works~\cite{eccv/UnlearnDiffAtk,icml/P4D,iclr/SalUn}.} Despite this lenient criterion, the detector demonstrated a low overall success rate, suggesting that the Nude Detector may not be fully reliable in reflecting the efficacy of nudity unlearning. And following~\cite{corr/SixCD}, we focus exclusively on six prominent nudity categories: "MALE\_BREAST\_EXPOSED", "MALE\_GENITALIA\_EXPOSED", "FEMALE\_BREAST\_EXPOSED", "FEMALE\_GENITALIA\_EXPOSED", "BUTTOCKS\_EXPOSED", and "ANUS\_EXPOSED".

\begin{table*}[ht]
\caption{Details of the constructed prompts for the \igmdata dataset. Each prompt targets specific unlearning task and includes varied descriptive elements.}
\label{tab:prompts}
\small
\centering
\begin{tabularx}{0.98\textwidth}{>{\centering\arraybackslash}m{0.08\textwidth} | >{\centering\arraybackslash}m{0.12\textwidth} | >{\centering\arraybackslash}m{0.2\textwidth} | >{\centering\arraybackslash}m{0.1\textwidth} | >{\centering\arraybackslash}m{0.18\textwidth} | >{\centering\arraybackslash}m{0.18\textwidth}}
\toprule
\textbf{Task} & \textbf{Target Words} & \textbf{Included Types} & \textbf{Replacement Candidates} & \textbf{Example $S \oplus t$} & \textbf{Example \ref{var_prompt:do_none} or \ref{var_prompt:do_opp}} \\ \midrule
Van Gogh  & Van Gogh style & thematic, landscape, still life, cityscape, dreamscape & - & A \colorfgt{Van Gogh style} picture of a sunflower in a vase.  &A picture of a sunflower in a vase.  \\ \midrule
Nudity   & naked, nude        & girl, boy, woman, man, person   & -    & A \colorfgt{nude} girl playing by the lake. &  A girl playing by the lake. \\ \midrule
Church    & church         & - &  airplane, bird, cat, chair, motorcycle, church        & A  \colorfgt{church} next to a river.   & A \textbf{bird} next to a river.  \\ \midrule
Parachute & parachute   & - &   airplane, bird, cat, chair, motorcycle, parachute       & \colorfgt{Parachute} drifting over a calm sea.  & \textbf{Airplane} drifting over a calm sea.    \\ \bottomrule
\end{tabularx}
\end{table*}

\section{Prompts and Cases}
\label{sec:app_benchmark}
We present the constructed prompts for each task in Table~\ref{tab:prompts}, which are used to generate the \textit{SD-GEN} portion of \igmdata. In addition, illustrative examples of generated images employed for benchmarking are shown in \figureautorefname{~\ref{fig:unlearning_case}}.

\begin{figure*}[ht]    
    \centering    
	\includegraphics[width=\textwidth]{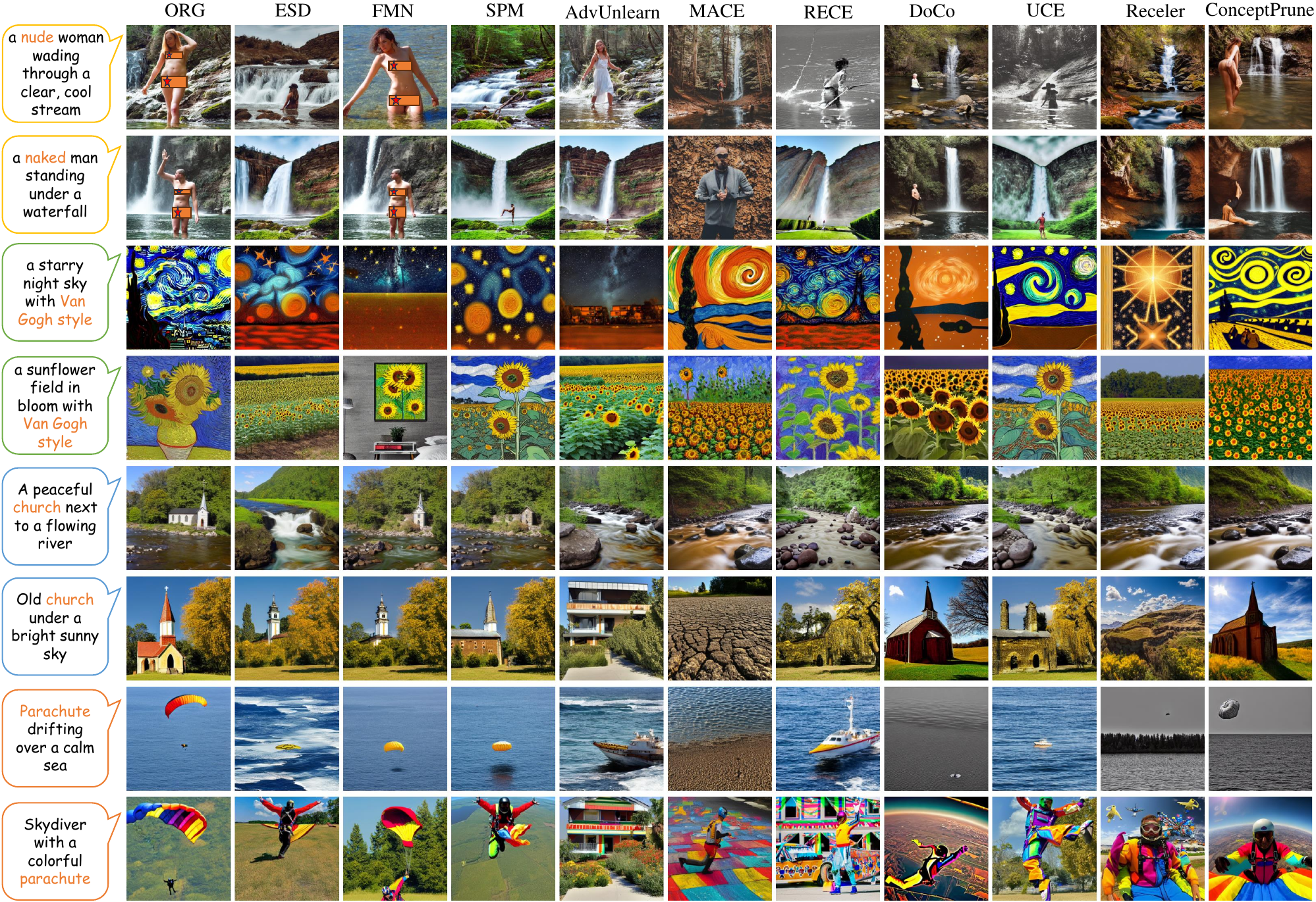}
	\caption{Examples of images generated by the original Stable Diffusion model and unlearned models for various unlearning tasks. The images are generated using the same prompts and random seeds for consistency across models.}
	\label{fig:unlearning_case}
\end{figure*}

\end{document}

%% file: dfn.tex
\newtheorem{observation}{\textbf{\textsc{Observation}}}

\newtheorem{lemma}{Lemma}

\definecolor{LightYellow}{RGB}{255,255,204}
\definecolor{LightBlue}{RGB}{204,229,255}
\definecolor{LightGreen}{RGB}{204,255,204}
\definecolor{LightPink}{RGB}{255,204,229}

\newcommand{\hide}[1]{}

\newcommand{\ml}{\mathcal}

\newlength\myindent
\newcommand{\bit}{\begin{compactitem}}
\newcommand{\eit}{\end{compactitem}}
\newcommand{\ben}{\begin{compactenum}}
\newcommand{\een}{\end{compactenum}}

\newcommand{\codeurl}{\url{https://github.com/ryliu68/IGMU}}

\newcommand{\colorfgt}[1]{\textcolor{brown}{#1}}

\newcommand{\model}{\ml{M}\xspace}

\newcommand{\igm}{\textsc{IGM}\xspace}
\newcommand{\igmutask}{\textsc{IGMU}\xspace}
\newcommand{\oframe}{\textsc{CatIGMU}\xspace}
\newcommand{\evalframe}{\textsc{EvalIGMU}\xspace}
\newcommand{\igmdata}{\textsc{DataIGM}\xspace}
\newcommand{\multiC}{\textit{MultClf}\xspace}

\newcommand{\wholeframe}{\textsc{IGMU}\xspace} 

\newcommand*\blkcircled[1]{
    \tikz[baseline=(char.base)]{
        \node[shape=circle,fill,inner sep=0.4pt](char){\textcolor{white}{#1}};}
}